\documentclass[default,iicol]{sn-jnl}

\usepackage{graphicx}%
\usepackage{multirow}%
\usepackage{amsmath,amssymb,amsfonts}%
\usepackage{amsthm}%
\usepackage{mathrsfs}%
\usepackage[title]{appendix}%
\usepackage{xcolor}%
\usepackage{textcomp}%
\usepackage{manyfoot}%
\usepackage{booktabs}%
\usepackage{algorithm}%
\usepackage{algorithmicx}%
\usepackage{algpseudocode}%
\usepackage{listings}%

\usepackage{epigraph}
\usepackage{url}            
\usepackage{nicefrac}       
\usepackage{colortbl}         
\usepackage{xspace}
\usepackage{mathtools}
\usepackage{overpic}
\usepackage{mathtools}
\usepackage{pifont}
\usepackage{mathtools}
\usepackage{float}
\usepackage{color}
\usepackage{rotating}
\usepackage{epsfig}
\usepackage{enumitem}
\usepackage{makecell}
\usepackage{lineno}
\usepackage{tabularx}
\usepackage{multirow}
\usepackage{hhline}
\usepackage{booktabs}
\usepackage{caption}
\usepackage{stmaryrd}
\usepackage[mathscr]{eucal}
\usepackage{lineno}
\usepackage[export]{adjustbox}

\usepackage{caption}
\usepackage{subcaption}

\usepackage{natbib} 

\theoremstyle{thmstyleone}%
\theoremstyle{thmstyletwo}%

\theoremstyle{thmstylethree}%

\definecolor{red}{rgb}{0.95,0.4,0.4}
\definecolor{purered}{rgb}{1,0,0}
\definecolor{blue}{rgb}{0.4,0.4,0.95}
\definecolor{darkblue}{rgb}{0,0,0.8}
\definecolor{darkred}{rgb}{1,0,0}
\definecolor{darkgreen}{rgb}{0,0.5,0}
\definecolor{darkergreen}{rgb}{0,0.5,0}

\definecolor{grey}{rgb}{0.6,0.6,0.6}
\definecolor{col1}{RGB}{232, 161, 148}
\definecolor{col2}{RGB}{148, 187, 232}
\definecolor{lightgrey}{rgb}{0.85,0.85,0.85}
\definecolor{lightlightgrey}{rgb}{0.9,0.9,0.9}
\definecolor{verylightBG}{rgb}{0.9,0.99,0.99}
\definecolor{darkgreen}{rgb}{0.3, 0.75, 0.3}

\newcommand\thc[1]{\textcolor{blue}{\bf #1}}
\newcommand\stc[1]{\textcolor{darkred}{\bf #1}}

\def\0{{\bf 0}}
\def\1{{\bf 1}}

\def\tha{\mbox{\boldmath$\theta$\unboldmath}}

\theoremstyle{plain}

\theoremstyle{definition}

\theoremstyle{remark}

\def\onedot{.\xspace}
\def\eg{\emph{e.g}\onedot} 
\def\ie{\emph{i.e}\onedot} 
\def\cf{\emph{c.f}\onedot} 
\def\etc{\emph{etc}\onedot} 
\def\wrt{w.r.t\onedot} 
\def\etal{\emph{et al}\onedot}

\usepackage[capitalize]{cleveref}
\crefname{section}{Sec.}{Secs.}
\Crefname{section}{Section}{Sections}
\Crefname{table}{Table}{Tables}
\crefname{table}{Tab.}{Tabs.}

\newcommand{\PAR}[1]{\vskip4pt \noindent {\bf #1~}}

\newcommand*{\unpub}[1]{\textcolor{gray}{\textbf{#1}}\@\xspace}

\newcommand*{\first}[1]{\textcolor{red}{\textbf{#1}}\@\xspace}

\newcommand*{\third}[1]{\textcolor{blue}{\textbf{#1}}\@\xspace}

\newcommand*{\thing}{\texttt{thing}\@\xspace}
\newcommand*{\stuff}{\texttt{stuff}\@\xspace}

\newcommand*{\other}{\texttt{other}\@\xspace}

\newcommand*{\known}{\texttt{known}\@\xspace}
\newcommand*{\unknown}{\texttt{unknown}\@\xspace}

\newcommand*{\lps}{\textit{LPS}\@\xspace}

\newcommand*{\hlps}{OWL\@\xspace}
\newcommand*{\hlpslong}{\textit{Open-World Lidar Panoptic Segmentor}\@\xspace}

\newcommand*{\gpkittilong}{Open-World Lidar Panoptic KITTI\@\xspace}

\newcommand*{\tp}{\text{TP}_c} 
\newcommand*{\fn}{\text{FN}_c} 
\newcommand*{\fp}{\text{FP}_c}

\definecolor{maroon}{cmyk}{0,0.87,0.68,0.32}

\definecolor{grey}{rgb}{0.6,0.6,0.6}

\definecolor{lightmaroon}{rgb}{0.98,0.86,0.86}
\newcommand{\pinkbgr}[1]{\colorbox{lightmaroon}{#1}}

\newcommand{\revision}[1]{\textcolor{black}{#1}}
\newcommand{\revisiontwo}[1]{\textcolor{black}{#1}}

\newcommand*{\markc}{\cellcolor{maroon!10}}

\begin{document}

\title[Article Title]{Lidar Panoptic Segmentation in an Open World}


\author*[1]{\fnm{Anirudh S} \sur{Chakravarthy}}\email{achakrav@andrew.cmu.edu}

\author[1]{\fnm{Meghana Reddy} \sur{Ganesina}}\email{mganesin@andrew.cmu.edu}

\author[1]{\fnm{Peiyun} \sur{Hu}}\email{peiyunh.ph@gmail.com}

\author[2]{\fnm{Laura} \sur{Leal-Taixé}}\email{leal.taixe@tum.de}

\author[3,4]{\fnm{Shu} \sur{Kong}}\email{skong@um.edu.mo}

\author[1]{\fnm{Deva} \sur{Ramanan}}\email{deva@andrew.cmu.edu}

\author[1]{\fnm{Aljosa} \sur{Osep}}\email{aosep@andrew.cmu.edu}

\affil[1]{\orgdiv{Robotics Institute}, \orgname{Carnegie Mellon University}} 


\affil[2]{\orgdiv{Dynamic Vision and Learning}, \orgname{TU Munich}}

\affil[3]{\orgdiv{Faculty of Science and Technology}, \orgname{University of Macau}}

\affil[4]{\orgdiv{Department of Computer Science and Engineering}, \orgname{Texas A\&M University}}

\abstract{
    Addressing Lidar Panoptic Segmentation (\lps) is crucial for safe deployment of autnomous vehicles. \lps aims to recognize and segment lidar points \wrt a pre-defined vocabulary of semantic classes, including \thing classes of countable objects (e.g., pedestrians and vehicles) and \stuff classes of amorphous regions (e.g., vegetation and road). Importantly, \lps requires segmenting individual \thing instances (\eg, every single vehicle).
    Current \lps methods make an unrealistic  assumption that the semantic class vocabulary is \textit{fixed} in the real open world, but in fact, class ontologies usually evolve over time as robots encounter instances of \textit{novel} classes \revision{that are considered to be unknowns \wrt thepre-defined  class vocabulary}. To address this unrealistic assumption, we study \lps in the Open World (LiPSOW): we train models on a dataset with a pre-defined semantic class vocabulary and \revision{study their generalization to a larger dataset} where novel instances of \thing and \stuff classes can appear. 
    This experimental setting leads to interesting conclusions. \revision{While prior art train class-specific instance segmentation methods and obtain state-of-the-art results on known classes, methods based on class-agnostic bottom-up grouping perform favorably on classes outside of the initial class vocabulary (\ie, unknown classes). Unfortunately, these methods do not perform on-par with fully data-driven methods on known classes. Our work suggests a middle ground: we perform class-agnostic point clustering and over-segment the input cloud in a hierarchical fashion, followed by binary point segment classification, akin to Region Proposal Network~\cite{Ren15NIPS}. We obtain the final point cloud segmentation by computing a cut in the weighted hierarchical tree of point segments, independently of semantic classification.
    %
    Remarkably, this unified approach} leads to strong performance \revision{on both known and unknown classes. 
    }
    %
}



\maketitle

\section{Introduction}
\label{sec:intro}

\begin{figure*}[t]
    \centering
    \includegraphics[width=0.99\linewidth]{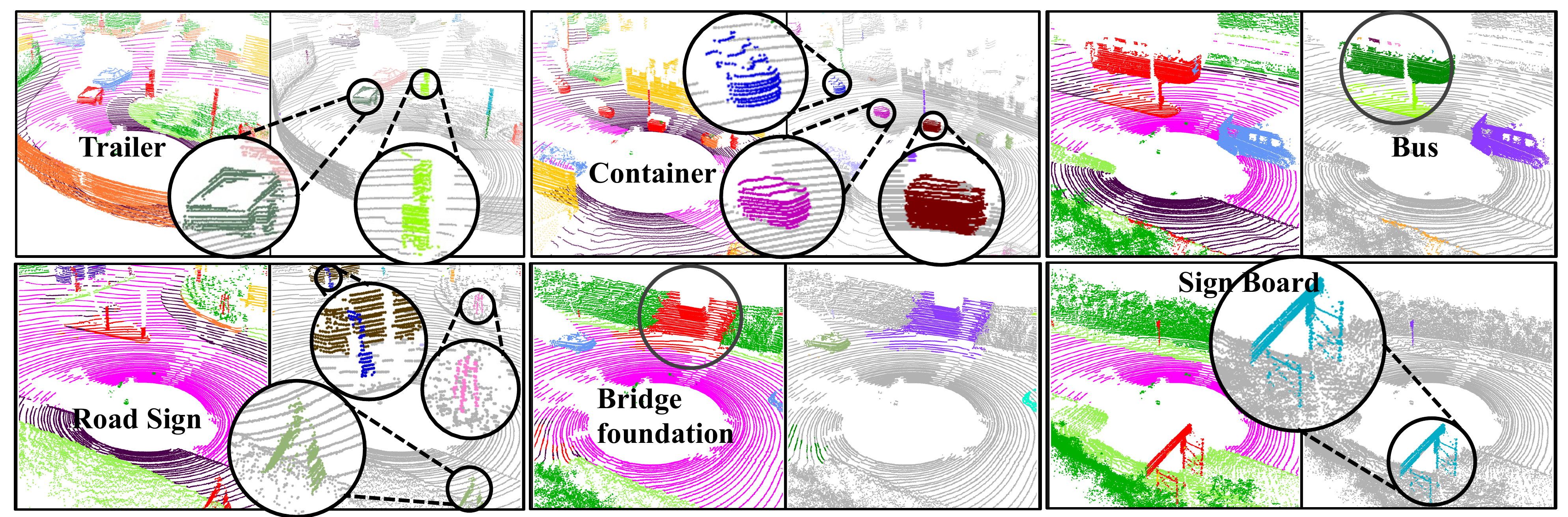}
    \caption{
    We study {\em Lidar Panoptic Segmentation} (\lps) in an Open World (\emph{LiPSOW}). 
    In each of the 2$\times$3 subfigure, left panel visualizes segmented points colored different w.r.t semantic classes, where \textcolor{red}{red} encodes \unknown; 
    right panel visualizes segmented \thing instances of \known and \unknown.
    For autonomous navigation, one should evaluate \lps methods in the presence of \emph{novel} {\tt thing} object instances and \stuff regions, which are usually termed as \unknown.
    We call this setting \emph{LiPSOW} setting, where
    methods should particularly segment points into \unknown object instances that are outside of 
    the $K$-way semantic classes in the predefined vocabulary.
    For example, given the predefined vocabulary by SemanticKITTI~\cite{Behley19ICCV}, the \unknown objects can be trailers, containers, signaling structures, highway bridge foundations, and buses, as visualized in this figure.
    }
    \label{fig:teaser}    
\end{figure*}

Lidar Panoptic Segmentation (LPS)~\cite{Behley21icra,fong21arxiv} unifies lidar point classification and segmentation, both important for autonomous agents to  interact with the open environment.
\revision{In LPS, each point must be classified as one of pre-defined $K$ semantic classes. LPS also defines the notion of \stuff and \thing: \thing classes are countable (\eg, \texttt{pedestrian} and \texttt{car} classes) and must be assigned unique instance identities. Amorphous regions, such as \texttt{vegetation} and \texttt{road}, are defined as \stuff.}
%
%
%
%



{\bf Motivation}. 
%
Solving \lps attracts increasing attention owing to its practical value for robotic applications, but the current setup fails to consider the realistic \emph{open-world} testing environments, where robots must know when they observe regions that do not fit in the predefined vocabulary of $K$ \revision{\known} classes (\eg, \texttt{fallen-tree-trunk} or \texttt{overturned-truck}) and recognize these regions as \revision{\unknown} obstacles. 
%

{\bf Lidar Panoptic Segmentation the Open World.} 
We study \emph{LPS in the open-world} (\emph{LiPSOW}, Fig.~\ref{fig:teaser}),
motivated by real-world challenges: AV companies have already operated autonomous fleets in different geo-locations, and these vehicles constantly observe new  or previously \unknown semantic classes over time. 
To study such situations, we introduce the LiPSOW evaluation protocol.
For example, LiPSOW allows one to train models on the SemanticKITTI~\cite{Behley19ICCV} by using its $K$ common  semantic classes for $K$-way classification of predicted segments, and importantly, \revision{gathering all the remaining rare classes as a catch-all \other class~\cite{hendrycks2018deep, kong2021opengan} to better detect \unknown objects}. 
Further, it performs evaluation on the KITTI360~\cite{Liao2021ARXIV} dataset, recorded in the same city with the same sensors but labeling more classes~\cite{lin2022continual}.
\revision{This effectively expands the class vocabulary to include instances of \unknown  classes.
The main challenge in \emph{LiPSOW} is to recognize and segment $K$ \known classes as defined in the SemanticKITTI vocabulary, and recognize \unknown  classes that appear in the testing set, i.e., KITTI360.} 

{\bf Technical Insights}.
Prior efforts in LPS~\cite{aygun21cvpr,gasperini2020panoster,hong2021lidar,li2022panoptic} \textit{learn} to group \texttt{known} classes but fail to generalize to \unknown classes. Based on this observation, Wong~\etal~\cite{wong2020identifying} suggests to learn to segment \texttt{known} classes and lean on bottom-up point clustering methods~\cite{Teichman11ICRA, nunes2022unsupervised,wong2020identifying,Moosmann09IVS} to segment \texttt{unknown} instances.
Our findings, derived from our  publicly available LiPSOW benchmark, suggest unified treatment of \texttt{known} and \texttt{unknown} classes: (i) we learn which points do \textit{not} correspond to $K$-known classes \revision{via outlier exposure}~\cite{hendrycks2018deep, kong2021opengan}, (ii) segment \unknown and \thing classes using class-agnostic bottom-up methods at multiple hierarchies, and (iii) learn which segments in the segmentation tree likely are objects by using labeled data, akin to class-agnostic training of Region Proposal Network~\cite{Ren15NIPS}. 
Surprisingly, this approach not only is effective for segmenting novel instances with high fidelity but also outperforms learned point-grouping-based methods for known \texttt{thing} classes, suggesting a \revision{unified treatment of \known and \unknown classes.}

{\bf Contributions}.
We make three major contributions: (i) We introduce LiPSOW, a new problem setting that extends \lps to the  \textit{open-world}, and establish an evaluation protocol to study LiPSOW. 
(ii) We repurpose existing \lps methods to address LiPSOW and comprehensively analyze their performance. 
(iii), drawing insights from in-depth analysis of existing \lps methods, we propose an approach that combines lidar semantic segmentation and a non-learned clustering algorithm with a learned scoring function. 
Our method effectively segments objects in a class-agnostic fashion, from both \known and \unknown classes during testing.
\revisiontwo{To foster future research, we make our \href{https://github.com/g-meghana-reddy/open-world-panoptic-segmentation}{code} publicly available. 
}

%

\section{Related Work}
\label{sec:relatedwork}

{\bf Lidar Semantic and Panoptic Segmentation.}
Recent Lidar Semantic Segmentation (LSS) and Lidar Panoptic Segmentation (LPS) methods are data-driven and fueled by the developments in learning representations from point sets~\cite{qi16cvpr,Qi17CVPR_pointnet,Thomas19ICCV} and publicly available densely labeled datasets~\cite{Behley19ICCV,Behley21icra, fong21arxiv}. 
LSS methods classify points into $K$ classes, for which dense supervision is available during training. Prior works focus on developing strong encoder-decoder-based architectures for sparse 3D \revision{data~\cite{qi16cvpr,Qi17CVPR_pointnet,Thomas19ICCV,yan2018second,choy20194d,tang2020spvnas,zhu2020cylindrical,loiseau2022online,ye2021drinet}}, the fusion of information obtained from different 3D \revision{representations~\cite{alonso20203d,xu2021rpvnet,li2022cpgnet}} or neural architecture search~\cite{tang2020spvnas}. 
On the other hand, LPS methods~\cite{Behley21icra,sirohi2021efficientlps,gasperini2020panoster} must additionally segment instances of \thing classes. 
Early methods combine Lidar semantic segmentation networks with 3D object detectors, with a heuristic fusion of both sources of information~\cite{Behley21icra}. Efficient-LPS~\cite{sirohi2021efficientlps} follows two-stage image-based object detection architectures using a range-image-based convolutional backbone. Several methods focus on end-to-end learning using \revision{point-based~\cite{Thomas19ICCV,hong2021lidar,hong2024unified,li2023center}} or sparse voxel~\cite{zhu2020cylindrical} backbones. 
\revision{Recent efforts focus on additional modalities such as camera-lidar fusion~\cite{marcuzzi2023ral,zhang2023lidar} or different views of Lidar data, such as range-view map~\cite{li2023cpseg} to improve model performance.} 
In addition to learning to classify points, these methods learn to group points in \revision{space~\cite{gasperini2020panoster,hong2021lidar,zhou2021panoptic,razani2021gp,li2022panoptic}}, space and time~\cite{aygun21cvpr,kreuzberg20224d}, or resort to bottom-up geometric clustering to segment instances~\cite{zhao2022divide, zhao2021technical}. 
Unlike our work, these methods do not consider the open-world environment, \revision{in which a pre-fixed class vocabulary is insufficient to capture all semantic classes that are encountered during the autonomous operation.}


{\bf Bottom-up Lidar Instance Segmentation.} Bottom-up grouping based on Euclidean distance has been used to isolate object instances in Lidar scans in a class-agnostic manner since the dawn of Lidar-based perception~\cite{thorpe1991toward}. Existing methods employ techniques such as flood-filling~\cite{douillard2011segmentation,Teichman11ICRA} and connected components~\cite{klasing2008clustering}, estimated in the rasterized bird's-eye view, bottom-up grouping \cite{Moosmann09IVS, Behley13IROS, Mcinnes2017OSS} using density-based clustering methods~\cite{Ester96KDD} or graph-based clustering methods \cite{Wang12ICRA}. 
Nunes~\etal~\cite{nunes2022unsupervised} propose to segment object instances with DBSCAN and refine segments using GraphCuts~\cite{boykov2006graph}. 
Since one-fits-all clustering parameters are difficult to obtain, Hu~\etal~\cite{hu2020learning} propose constructing a hierarchical tree of several plausible Lidar segmentations, obtained using a density-based clustering method~\cite{Ester96KDD}. These regions are then scored using a learned objectness regressor, and optimal instance segmentation (\wrt the regressed objectness function) can \revision{then be obtained via a cut in this tree.} In this paper, we demonstrate that a data-driven Lidar Panoptic Segmentation network, in conjunction with hierarchical tree construction, forms a strong baseline for Lidar Panoptic Segmentation in an Open World.


{\bf Domain Adaption for Lidar Segmentation.} 
\revision{Domain adaptation aims to improve the generalization ability of segmentation models, trained on the source domain, by adapting models to the (unlabeled) target domain. Prior works focus on adapting 3D representations~\cite{langer2020domain,yi2021complete}, feature representations~\cite{rist2019cross,jiang2021lidarnet,wu2019squeezesegv2,shaban2023lidar, kong2023conda}. While our paper focuses on identifying unseen novel objects (\unknown's) under similar sensor distributions and geographical regions, these methods focus on adapting to target distributions under significant sensor configuration shifts or environments.}

{\bf Open-Set Recognition (OSR)} requires training on data from $K$ known classes and recognizing examples from unknown classes encountered during testing~\cite{scheirer2012toward}. 
Many OSR approaches train a $K$-way classification network and then exploit the trained model for OSR~\cite{yoshihashi19cvpr, OzaP19}. 
Recent work shows that a more realistic setup is to allow access to some outlier examples during training~\cite{hendrycks2018deep, kong2021opengan}. Such outliers are used as instances of {\tt other} class \revision{(\ie, held-out samples that do not correspond to pre-defined K-classes)} during training, significantly boosting OSR performance. In the context of (lidar) semantic segmentation, \cite{kong2021opengan, cen2022open} approximate the distribution of novel objects by synthesizing instances of novel classes. 
Different from the aforementioned, we tackle OSR through the lens of lidar panoptic segmentation.

\revision{
{\bf Open-Vocabulary Object Detection.}
Recent efforts utilize such bottom-up segmentation, combined with Kalman-filter-based object trackers (\cf, \cite{Teichman12IJRR, Dewan15ICRA, Osep18ICRA, Osep20ICRA}) to pseudo-label instances of moving objects in Lidar~\cite{najibi2022motion,zhang2023towards} or stereo video streams~\cite{Osep19ICRA,osep18ECCVW} and use these instances to train object detectors for moving object instances. Moreover, \cite{najibi2023unsupervised} demonstrate that detected moving objects can also be classified in a zero-shot manner by distilling CLIP~\cite{radford2021learning} features to Lidar.
Different from the aforementioned, we tackle \lps, which entails dense segmentation and recognition of \thing and \stuff classes for moving, as well as stationary objects. Segmented instances that our method classifies as \unknown class could be further classified in a similar fashion as proposed in \cite{najibi2023unsupervised} to obtain a fine-grained semantic interpretation of segmented regions.
%
}


{\bf Open-set (Lidar) Segmentation.} 
Early works by~\cite{Teichman11ICRA}, \cite{Moosmann09IVS, Moosmann13ICRA} and \cite{Held16RSS} can be understood as early attempts towards open-set Lidar instance segmentation. In \cite{Teichman11ICRA} after bottom-up segmentation of individual point clouds, objects are tracked across time, and classified as \texttt{car}, \texttt{cyclist}, \texttt{pedestrian} or \texttt{other}. 
The most similar works to ours are \cite{wong2020identifying,hwang21CVPR}, which study OSR in lidar point clouds and images, respectively. Their setup assumes complete annotations for {\tt stuff} classes, \ie, assume \stuff classes exhaustively labeled. This is an unrealistic assumption since new \stuff classes (\eg, bridges and tunnels) may also be encountered at test time and must be recognized as novel classes. 
\revision{Differently, our setup assumes novel classes (\ie, \unknown's classes) can appear in both \stuff and \thing classes.}  
This is a realistic setup, further justified by the ontology change from SemanticKITTI to KITTI360~\cite{lin2022continual} where several new \stuff classes are encountered (\eg, Fig.~\ref{fig:cross-kitti-viz}: gate, wall, tunnel, bridge, garage, stop, rail track). This subtle yet crucial distinction separates LiPSOW from previous settings.
%
Prior works~\cite{wong2020identifying,hwang21CVPR,cen2022open} build their methods and evaluation on the assumption that the unknown consists of only \thing, \ie, assuming complete annotations for \stuff. Our experimental validation (Sec.~\ref{sec:experiments:glps}) confirms that the open-set semantic segmentation method~\cite{cen2022open}, which assumes complete annotation for \stuff classes, does not perform well in our proposed setup. While \cite{wong2020identifying} is the first work investigating \lps in open-set conditions, it conducts experiments on a proprietary dataset, \revision{and does not release the code or data}. We repurpose publicly available datasets to foster future research on LiPSOW. Finally, we suggest a different approach for LiPSOW that unifies instance segmentation of \known and \unknown classes in a class-agnostic manner, by contrast to \cite{wong2020identifying} that learns to segment \known classes and \revision{\textit{only}} segment instances of novel classes via DBSCAN.  

\section{Open World Lidar Panoptic Segmentation}
\label{sec:glps}
In this section, we review the problem of Lidar Panoptic Segmentation (\lps) and discuss the limitations of its setup from the perspective of open-world deployment in Sec.~\ref{sec:lps}.
To address the limitations, we introduce \lps in an Open World setting (LiPSOW) in Sec.~\ref{sec:lipsow}. 

\subsection{Lidar Panoptic Segmentation}
\label{sec:lps}

{\bf Definition}.
\lps  takes a Lidar point cloud $\mathcal{P} = \{ \mathbf{p}_i \in \mathbb{R}^3\}_{i=1}^N$ as input, and aims to classify points \wrt a predefined vocabulary $\mathcal{K} = \{1, \ldots, K\}$ of $K$ categorical labels and segment object instances. Categorical labels are divided into (1) {\tt thing} classes, covering countable objects such as cars and persons, and (2) \stuff classes that cover uncountable amorphous regions such as road and vegetation. For \thing points, \lps methods must segment object instances (\eg, every car). 
Mathematically, \lps methods learn a function $f(\cdot; \tha)$ parameterized by $\tha$, mapping an input point $\mathbf{p}_i$ to a semantic label $k$ and object instance ID$_i$, \ie,
$f(\mathbf{p}_i; \tha) \to (k, \text{ID}_i)$, where $k\in\mathcal{K}$, ID$_i$ is a unique ID for the object instance $\mathbf{p}_i$ belongs to, and particularly, ID$_i=0$ means that $\mathbf{p}_i$ belongs to one of the \stuff classes (\ie, class-$k$ is \stuff). 
\lps measures the per-point classification accuracy (\ie, Lidar semantic segmentation) and per-instance segmentation accuracy.

{\bf Remarks}.
\lps does not properly formulate the real-world case that there exist points belonging to an \revision{{\tt unknown} catch-all superclass}, which contain various unknown classes encountered only during testing. 
For example, a vocabulary in AVs probably does not have labels such as {\tt sliding}-{\tt unattended}-{\tt stroller} and {\tt fallen}-{\tt tree}-{\tt trunk}, but AVs must segment them into individual instances for safe maneuver such as ``stop'', ``yield'' or ``change-lane''.
We are motivated to address this in detail below.

\subsection{LPS in an Open World} 
\label{sec:lipsow}

{\bf Definition}.
Extending \lps, {\em Lidar Panoptic Segmentation in the Open World} (LiPSOW) further requires classifying points into an {\tt unknown} class if the points do not belong to any of the predefined $K$ semantic classes, and segment them as {\tt unknown} instances. 
That said, \revision{{\tt unknown}} covers all classes that do not correspond to any of the $K$ predefined classes and might contain unannotated instances that, without prior knowledge, cannot be treated as a \stuff or \thing class. 
Formally, we define the \revision{{\tt unknown}} class as the $(K+1)^{th}$ class, so LiPSOW methods learn a function $f(\cdot; \tha)$ parameterized by $\tha$, mapping input points $\mathbf{p}_i$ to a semantic label $k$ and instance ID$_i$, \ie,
$f(\mathbf{p}_i; \tha) \to (k, \text{ID}_i)$, where $k\in \{1, \dots, K, K+1\}$. As before, ID$_i$ is a unique ID for point-$i$ and ID$_i=0$ implies that class-$k$ is \stuff. 

{\revision{\bf Significance.}}
By definition, LiPSOW algorithms should be able to distinguish \revision{{\tt unknown}} objects from the predefined $K$ classes and to segment corresponding object instances.
This ability is useful for many applications. 
First, recognizing unknown objects is crucial for safety-critical robotic operations, \eg, AVs should recognize a never-before-seen sliding-unattended-stroller to avoid collision and casualty. 
Second, unknown instances could be used in conjunction with active learning~\cite{ren2021survey,zhan2022comparative} to help select more valuable examples that are recognized as unknown to reduce data collection and annotation costs. 
 
{\bf Remark I: Can \revision{\tt unknowns} be seen during training?}
Related to LiPSOW is open-set recognition, the task of recognizing \revision{\tt unknown} examples during test-time. A conventional setup is that all the training data is labeled \wrt the $K$ predefined classes, and test-time examples \revision{may originate from any of the $K$ \known classes, or the ($K$+1)$^{th}$ {\tt unknown}~\cite{scheirer2012toward, bendale2016towards, yoshihashi2019classification, OzaP19} class}. While these works suggest that \revision{{\tt unknown}} examples should not be part of the training set, recent work has comprehensively demonstrated that a more reasonable setup is to explicitly exploit outlier data or diverse {\tt other} examples during training. In particular, \cite{hendrycks2018deep, kong2021opengan} show that such models effectively generalize to real unseen examples. 
To improve the real-world AV application, we consider the latter setup, \ie, \revision{the training set contains instances of \known classes, labeled as {\tt other}. Instances of these classes are available during the training, but, importantly, do not overlap with the $K$ \known classes. These classes are presented as possible instances of the \unknown, ($K$+1)$^{th}$ class.} Note that the {\tt other} class is often named {\tt void} or {\tt unlabeled} in many contemporary benchmarks~\cite{cordts2016cityscapes, neuhold2017mapillary}. 

{\bf Remark II: How to define \revision{{\tt unknown}} semantic classes?} 
Strictly defining all objects that could appear in the open-world would be difficult. Instead, inspired by the image segmentation literature~\cite{martin2001database,fomenko2022learning}, which suggests that humans have a consensus on which image regions constitute object instances, we define \revision{\unknown} instances as those that were (i) not annotated \wrt a categorical label in the train-set but
(ii) labeled as objects by human annotators in the test-set. 

{\bf Remark III. Can only novel \thing classes be regarded as \unknown's?}
Prior efforts tackling open-set in images and point clouds \cite{wong2020identifying,hwang21CVPR} assume \stuff classes exhaustively labeled. This is an unrealistic assumption since new \stuff classes (\eg, Fig.~\ref{fig:cross-kitti-viz}: tunnel, bridge, rail track, etc.) may also be encountered at test time and must be recognized as novel classes. This subtle yet crucial distinction separates LiPSOW from prior work. %


\section{Methodology}
\label{sec:hlps}
\begin{figure*}[t]
  \centering
  \includegraphics[width=0.99\linewidth]{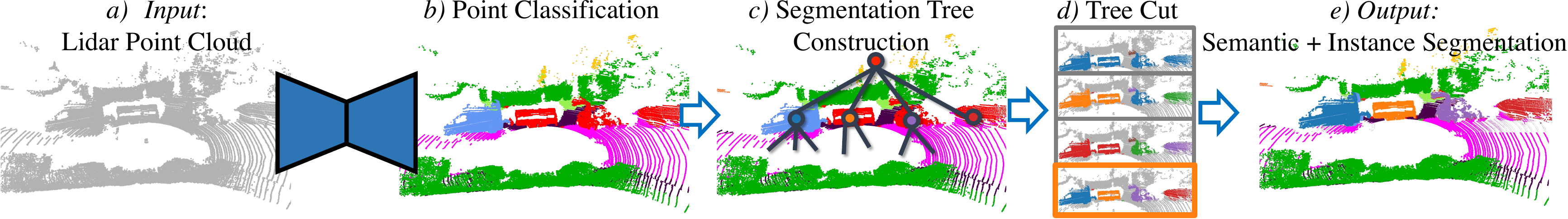}
  \caption{\small \textbf{Open-World Lidar Panoptic Segmentation \revision{(\hlps)}}: We first perform $K+1$ semantic segmentation network on a point cloud \textit{(a--b)} and classify points as {\tt stuff}, {\tt things}, and \textcolor{red}{\unknown} (\revision{point color encodes semantic classes, \textcolor{red}{red} points represent \unknown's)}. 
  Then we construct a hierarchical tree of ``all possible'' segments for {\tt thing} and \textcolor{red}{\unknown} points (\textit{c}) and train a segment-scoring function to cut the tree (\textit{d}), finally producing instance and semantic segmentation results (\textit{e}). 
  }
  \label{fig:hlps}
\end{figure*}

Existing methods for \lps~\cite{aygun21cvpr,gasperini2020panoster,zhou2021panoptic,razani2021gp} learn to classify points, and {\it learn} to group points that represent \thing classes. These methods work under the \lps setting, where semantic and instance-level supervisions are given for all classes. 
For LiPSOW, we need to rethink existing methodology: LiPSOW methods must (just as in \lps) recognize predefined semantic classes and segment instances of \thing classes. Additionally, they need to cope with the inherent difficulty of recognizing and segmenting the \revision{\unknown} class that \textit{mixes} \stuff and 
{\tt things}. Unfortunately, for this catch-all class, exhaustive semantic and instance-level supervision is not available. 

To design a LiPSOW method, named \hlpslong (\hlps), we draw inspiration from two-stage object detectors~\cite{Ren15NIPS}. It was shown in literature that such networks can be repurposed for image-based open-set~\cite{dhamija18boult} and open-world object detection~\cite{hwang21CVPR,weng21cvpr,liu2022opening,joseph21CVPR,fomenko2022learning}. 
Two-stage networks were also adopted for 3D object detection~\cite{Shi19CVPR,Chen15NIPS}. However, a 3D analogy of the region proposal network (RPN), a key component that allows us to recognize instances of novel classes, is not trivial due to the large 3D search space~\cite{Shi19CVPR,Chen15NIPS}. For this reason, prior works constrain the set of anchor boxes to the mean size of each semantic class (\eg, \texttt{car} and \texttt{pedestrian} sized boxes). This approach is not scalable to the large variety of object instances that may appear in the \other class. Instead, we rely on the observations of early work on Lidar perception~\cite{Teichman11ICRA,Teichman12IJRR,Held16RSS,Behley13IROS}, which shows that simple bottom-up grouping of points yields a compact set of class-agnostic object candidates. In the following sections, we outline a simple and effective method for LiPSOW, based on data-driven approaches~\cite{Thomas19ICCV,aygun21cvpr} and perceptual grouping~\cite{klasing2008clustering,douillard2011segmentation,hu2020learning,Behley13IROS} based methods, as well as recent findings in open-set recognition~\cite{kong2021opengan}.

{\bf High-level overview:}
\revision{We propose a two-stage network for LiPSOW, which is trained sequentially.}
We adopt an encoder-decoder point-based backbone~\cite{Thomas19ICCV} to learn to classify points in $K+1$ fashion. We explicitly train our network to distinguish points from $K$ \known classes from the \other class, \revision{that is considered to be a representative of the \unknown class during the model training} (Fig.~\ref{fig:hlps}\textit{b}). This network estimates label predictions that belong to {\tt stuff}, {\tt things}, and the mixed {\other} class.  
 In the second stage, we run a non-learned  clustering algorithm on points recognized as {\tt thing} or {\tt other} (Fig.~\ref{fig:hlps}\textit{c}) and apply a learned scoring function to derive the final instance segmentation. To this end, we produce a hierarchical segmentation tree (\cf, \cite{hu2020learning}), and train the second-stage network that learns to estimate how likely a point segment is an object, and run a min-cut algorithm~\cite{hu2020learning} to obtain a unique, globally optimal point-to-instance assignment (Fig.~\ref{fig:hlps}\textit{d}). Importantly, this method treats the \thing and \other classes in a unified manner, producing instance segmentation for both. We present individual components of our \hlpslong (\hlps) baseline below. 

{\bf Semantic segmentation network.} 
We train an encoder-decoder architecture that operates directly on point cloud $P \in \mathbb{R}^{N\times 3}$. In particular, we train a well-consolidated KPConv~\cite{Thomas19ICCV} network with deformable convolutions; however, we note that a variety of backbones suitable for learning representations from unstructured point sets could be used~\cite{qi16cvpr,yan2018second}.  
\revision{We use the KPConv-based LPS network~\cite{aygun21cvpr} due to its (i) open-source implementation and (ii) point-based backbone, that directly learns fine-grained per-point features, as opposed to 3D sparse-convolutional networks~\cite{choy20194d} that estimate per-voxel features.} 
We attach a semantic classifier on top of the decoder feature representation $F \in \mathbb{R}^{N \times D}$ that outputs a semantic map $S \in \mathbb{R}^{N \times {(K + 1)}}$. Finally, we train this network head using the cross-entropy loss. 
The difference from conventional (lidar) semantic segmentation training is that we explicitly introduce an additional catch-all class by holding out rare  \revision{(\other)} classes during the model training (Sec~\ref{para:vocab}). This class is analogous to the catch-all \textit{background} class~\cite{Ren15NIPS}, a common practice in training two-stage object detectors. The final ($K$+1)-way softmax provides a smooth distribution that indicates the likelihood of a point being one of $K$ classes or the \other class. \revision{Classes classified as \other during test time are considered to be \unknown's.}

\revision{Such a catch-all \other class is not common practice in training semantic segmentation networks, as it is usually assumed that points (or pixels) are densely and exhaustively labeled. However, in LiPSOW, this assumption is no longer valid. Without it, we would incentivize the network to label each point as one of the $K$ classes.} 
%




\revision{{\bf Segmenting any object.}}  Using a proximity-based point grouping method, we can construct combinatorially many possible point segments from a point cloud of size $N$. We learn a function $f(p) \to [0,1],\;p \subset P \in \mathbb{R}^{N\times 3}$ that scores \textit{objectness} of a subset of points in a data-driven manner to indicate \textit{how likely} a point segment encapsulates an object. This is analogous to image-based object proposal generation methods~\cite{Alexe12TPAMI,Zitnick14ECCV} that adopt sliding window search and learn an objectness score to rank windows. The advantage of such methods over recent work on data-driven pixel/point grouping~\cite{kong2018recurrent, aygun21cvpr,razani2021gp,zhou2021panoptic} is that the set of possible objects should naturally cover all objects, irrespective of class labels. 

\revision{To understand if our segmentation tree covers most of the relevant objects, we measure recall using labeled instances.} To this end, we follow~\cite{hu2020learning} and construct a hierarchical segmentation tree $T$ by applying Euclidean clustering recursively with decreasing distance threshold using DBSCAN~\cite{Ester96KDD} and parameters recommended by~\cite{hu2020learning}. \revision{Our experiments show that with this approach we can recall $97.2\%$ of instances labeled in the SemanticKITTI dataset~\cite{Behley19ICCV,Behley21icra} validation set (see Sec.~\ref{sec:experiments:pls}).}
This shows that not only can this approach segment a large variety of objects, but also it does not need to learn how to group instances of \known classes. \textit{These instances are already included in the segmentation tree.}
    
\revision{{\bf Learning an objectness function.}} 
There are several ways to learn such a function $f(p) \to [0,1]$ that estimates how likely a subset of points represents an object. 
One approach is to estimate a per-point objectness score. Following~\cite{aygun21cvpr}, this can be learned by regressing a truncated distance $O \in \mathbb{R}^{N \times 1}$ to the nearest point center of a labeled instance~\cite{aygun21cvpr} atop of decoder features $F \in \mathbb{R}^{N \times D}$. The objectness value can then be averaged over the segment $p \subset P$. 
Alternatively, we can train a holistic classifier as a second-stage network by pooling point segment features, followed by fully-connected layers, similar to the PointNet~\cite{qi16cvpr} classification network. 
In this case, we pre-built hierarchical segmentation trees $T_i$ for each point cloud $i$ in the training set and minimize the training loss based on the signal we obtain from \textit{matched} segments between the segmentation trees and set of labeled instances, $GT_i$. 
One possibility is to use binary cross-entropy loss (similar to how the RPN is trained); alternatively, we can directly regress the objectness value to be proportional to the point-set intersection-over-union.
We detail the network architecture and training recipes in the appendix (Sec.~\ref{sec:sup:implementation}) and discuss design choices on how to train such a network in Sec.~\ref{sec:experiments:pls}. 

\revision{{\bf Unique point-to-instance assignment.}}
Segmentation tree $T$ provides a hierarchy of pairs of point segments and their corresponding scores. However, for LiPSOW, we need to assign points to instances uniquely. Intuitively, this property will be satisfied with any ``cut'' in this tree; we could simply output leaf nodes in a tree after the cut is performed. The question then boils down to \textit{where} to cut such that the overall segmentation score is as good as possible according to some criterion. 
It was shown in~\cite{hu2020learning} that we can compute optimal worst-case segmentation efficiently by simply traversing the tree, as long as we have \textit{strictly smaller} segments at each tree-level. Optimal worse-case segmentation is the segmentation that yields an overall (global) segmentation score when the overall segmentation score is defined as the \textit{worst} objectness among its local segments (this can be efficiently evaluated by looking at the tree leaf nodes).
\revision{This approach ensures a unique point-to-segment assignment \ie no overlap.}
This algorithm is not the contribution of this paper. However, for completeness, we provide the algorithm with a detailed explanation in the appendix (\revision{Sec.~\ref{sec:sup:tree_gen}}). 

{\bf Inference.} At inference time, we first make a forward pass through our network, construct the segmentation tree on points classified as \thing or \other (\revision{\ie \unknown}) class, and run our objectness classifier for each segment in the tree. After such construction, we run the tree-cut algorithm to obtain unique point-to-instance assignments. Finally, as semantic labels within segments may be inconsistent, we assign a majority vote to each segmented instance.

\section{Experiments}
\label{sec:experiments}
In this section, we first outline our experimental setup for LiPSOW (Sec~\ref{sec:glps_eval}). Then, we discuss and ablate our \hlpslong (\hlps) (Sec.~\ref{sec:experiments:pls}) on a standard lidar panoptic segmentation (\lps) benchmark, SemanticKITTI~\cite{Behley19ICCV,Behley21icra}. Finally, we demonstrate the generality of our approach with cross-dataset evaluation (Sec~\ref{sec:experiments:glps}).

\subsection{Evaluation Protocol of LiPSOW} 
\label{sec:glps_eval}

We set up an evaluation protocol to simulate the conditions that occur when a robot within a certain geographic region (\ie, city), \eg, a robot taxi fleet. 
Here, data is recorded by the same sensor type (cross-sensor domain generalization is out-of-scope of this paper), which is reasonable in practice when deploying robot taxis of the same type. 
Importantly, however, even though we focus on a certain geographic region, \textit{source} and \textit{target} data must be recorded at disjunct locations (\ie, sequences that appear in the test set should not be recorded in precisely same city districts). In such a setting, domain shifts are often gradual \eg, we record more data over time, and therefore observe a larger variety of \known regions and more \other objects that appear in the long-tail of the object class distribution.

\begin{figure*}[ht]
     \centering
     \begin{subfigure}[ht]{0.47\linewidth}
        \centering
        \resizebox{0.99\linewidth}{!}{%
            \begin{tabular}{lll}
            \toprule 
            \multicolumn{1}{c}{} & SemanticKITTI & \multicolumn{1}{c}{KITTI360}\tabularnewline
            \hline 
            Sensor & Velodyne HDL-64E \quad & Velodyne HDL-64E\quad\quad \tabularnewline
            Geographic region \quad\quad & Karlsruhe, GER & Karlsruhe, GER\tabularnewline
            Distance & 39.2 km & 73.7 km\tabularnewline
            Recording date & 2011 & 2013\tabularnewline
            \# Scans & 23k & 80k\tabularnewline
            \# Classes & 21 & 37 \tabularnewline
            \# Instance classes  \quad& 8 & 24\tabularnewline
            \bottomrule 
            \end{tabular}
        }
         \caption{\textbf{SemanticKITTI and KITTI360 datasets.} Both were recorded with the same sensory setup (in 2011 and 2013) in the same city but in different, non-overlapping city districts. KITTI360 contains a significantly larger number of classes with instance labels.}
         \label{tab:cross-pan}
     \end{subfigure}
     \hfill
     \begin{subfigure}[ht]{0.49\linewidth}
         \centering
         \includegraphics[width=0.99\linewidth]{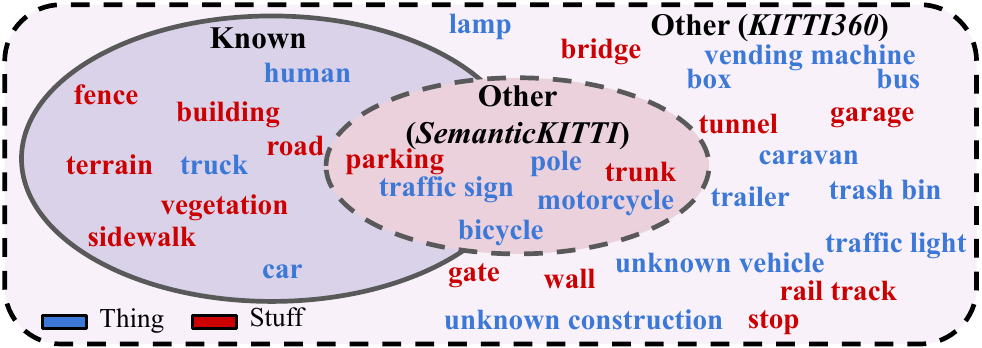}
         \caption{
            \textbf{Open-World Setting (\revision{Vocabulary 1})}: Source domain class-vocabulary (defined in SemanticKITTI) is split into \known and \other classes. KITTI360 is a super-set vocabulary that extends the \known and \other class with a larger number of classes, labeled only in KITTI360.
         }
         \label{fig:classdef}
     \end{subfigure}
     \caption{We base LiPSOW setup on SemanticKITTI~\cite{Behley19ICCV,Behley21icra} and KITTI360~\cite{Liao2021ARXIV} datasets. We train and validate models using SemanticKITTI, and re-purpose KITTI360 dataset, which contains classes, not labeled in SemanticKITTI (\ie, \unknown) as a test set. \revision{A detailed list of our taxonomy is provided in the appendix (Table ~\ref{tab:sup:task_split}).}}
     \label{fig:cross-kitti-viz}
\end{figure*}

{\bf \gpkittilong.}
To study LiPSOW in such a setting, we base our experimental setup on SemanticKITTI~\cite{Behley19ICCV} and KITTI360~\cite{Liao2021ARXIV} datasets. They were recorded in \textit{distinct} regions of Karlsruhe, Germany (Tab~\ref{tab:cross-pan}). We use SemanticKITTI for model training and validation, and KITTI360 sequences \textit{only} for testing. 

{\bf Source (train/val) domain.} \label{para:vocab} In Fig.~\ref{fig:classdef} (blue ``known'' set) we visualize the SemanticKITTI classes~\cite{Behley19ICCV}. 
%
%
The dashed inner circle (pink) denotes the rarer classes, which we merge into a single \other class. 
Those are examples of regions, different from $K$ known classes, \ie, $(K+1)^{th}$ catch-all class. This allows evaluating how well the model learns to separate \known classes from \other by measuring IoU for the \other class and mIoU for all classes within the source domain. 
\revision{We call this taxonomy Vocabulary 1.}
\revision{
However, by this construction of Vocabulary 1, classes such as \texttt{bicycle} and \texttt{motorcycle}, which belong to \other, are important for autonomous driving and commonly observed in urban environments. 
Therefore, we also construct Vocabulary 2, which holds out only the rarest categories as \other. We provide further details on vocabulary construction and taxonomy in the appendix (Sec.~\ref{sec:sup:lipsow} \revision{and Table~\ref{tab:sup:task_split}}).
}

{\bf Target (test) domain.} We evaluate models on KITTI360, which encompasses all SemanticKITTI classes, and \emph{importantly}, additional $10$ \thing (with instance labels) and $7$ \stuff classes, which are used as novel classes in experiments. We discuss vocabulary changes that ensure SemanticKITTI and KITTI360 vocabularies are consistent in the appendix (Sec.~\ref{sec:sup:lipsow} and Table~\ref{tab:sup:task_split}).

{\bf Metrics.} 
We repurpose evaluation metrics proposed in the context of semantic segmentation (mean intersection-over-union, mIoU~\cite{Everingham10IJCV}) and panoptic segmentation (panoptic quality, PQ~\cite{Kirillov19CVPR}). 
To quantify the point classification performance (mIoU), we simply treat \other class as ``just one more class''. 
To quantify panoptic segmentation, we split the evaluation for \known classes and \other classes. We evaluate \known using $PQ = SQ \times RQ$, as defined by~\cite{Kirillov19CVPR}. The segmentation ($SQ$) term averages instance-level IoU for each true positive (TP), while the recognition quality ($RQ = \frac{\tp}{|\tp| + \frac{1}{2} |\fp| + \frac{1}{2} |\fn|}$) is evaluated as F1 score (harmonic mean of precision and recall).  
For \other classes, the task definition does not specify which semantic classes are \thing classes, nor specifies the vocabulary of target instance classes. As we cannot annotate every possible semantic class, it is important to not penalize \textit{false positives}, FPs, as these cannot be clearly defined. Therefore, we follow~\cite{wong2020identifying,liu2022opening}, and replace the $RQ$ term with recall ($\frac{\tp}{|\tp| + |\fn|}$), which we call UQ (unknown quality). Note that our UQ is computed slightly differently from the UQ introduced by~\cite{wong2020identifying} because we do not penalize segment predictions that overlap unlabeled {\tt stuff} in the {\tt other} class.

\subsection{\revision{Lidar Panoptic Segmentation}}
\label{sec:experiments:pls}

\revision{A pre-requisite for good performance on LIPSOW is good performance on \known classes. Therefore, first, we present a comparison of our method with state-of-the-art \lps methods on SemanticKITTI~\cite{Behley19ICCV} in Tab.~\ref{tab:bench_semkitti}.
}
The top performing method on the val-set for \known classes is \textit{GP-S3Net}~\cite{razani2021gp}. Due to its strong Transformer-based backbone, \textit{GP-S3Net} obtains $73.0\%$ mIoU and the highest PQ of $63.3\%$.
The remaining methods are in the ballpark of $60-65\%$ mIoU and $55-59\%$ PQ. \revision{We note that methods such as \textit{CPGNet}~\cite{li2022cpgnet} and \textit{LCPS}~\cite{zhang2023lidar} are also highly performant, however, these utilize multi-modal inputs (range-views and images, respectively). We base our \hlps on well-consolidated and easily-extendable KPConv~\cite{Thomas19ICCV}, similar to 4D-PLS.}
%
\footnote{GP-S3Net~\cite{razani2021gp}, \revision{DSNetv2~\cite{hong2024unified}, Panoptic-PHNet~\cite{li2022panoptic}} do not provide source code.}

\begin{table*}[ht]
    \centering
    \small 
    \setlength{\tabcolsep}{3pt}
    \caption{\textbf{\lps results on SemanticKITTI validation set.} Alhough LPS is not our primary focus, we show ``closed-world" panoptic accuracy, \revision{since good performance on \known classes is a pre-requisite for LiPSOW}.
    Our method is competitive with state-of-the-art methods that adopt stronger semantic backbones.
    Given GT semantic labels, our method achieves nearly perfect PQ ($98.3\%$), indicating that future efforts should focus on semantic classification. 
    }
    \resizebox{1.0\linewidth}{!}{%
    \begin{tabular}{cl|c|ccc|ccc|ccc|c|cc}
    \toprule
    & Method & PQ & PQ\textsuperscript{\textdagger} & RQ & SQ & PQ\textsuperscript{Th} & RQ\textsuperscript{Th} & SQ\textsuperscript{Th} & PQ\textsuperscript{St} & RQ\textsuperscript{St} & SQ\textsuperscript{St} & 
    mIoU & Prec\textsuperscript{Th} & Rec\textsuperscript{Th}\\
    \midrule
    \parbox[t]{2mm}{\multirow{12}{*}{\rotatebox{90}{Baselines}}} & Panoster~\cite{gasperini2020panoster} & 55.6 & -- & 66.8 & {79.9} & 56.6 & 65.8 & -- & -- & -- & -- & 61.1 & -- & -- \\ 
    & \revision{DS-Net}~\cite{hong2021lidar} & 57.7 & 63.4 & 68.0 & 77.6 & 61.8 & 68.8 & 78.2 & 54.8 & 67.3 & 77.1 & 63.5 & -- & --  \\
    & \revision{DS-Net v2}~\cite{hong2024unified} & 61.4 & 65.2 & 72.7 & 79.0 & 65.2 & 72.3 & 79.3 & 57.9 & 71.1 &  79.3 & 69.6 & -- & --  \\
    & PolarSeg-Panoptic~\cite{zhou2021panoptic} & 59.1 & 64.1 & 70.2 & 78.3 & {65.7} & {74.7} & {87.4} & 54.3 & 66.9 & 71.6 & 64.5 & -- & -- \\
    %
    & MaskPLS~\cite{marcuzzi2023ral} & 59.8 & -- & 69.0 & 76.3 & -- & -- & -- & -- & -- & -- & 61.9 & -- & -- \\
    & Efficient-LPS~\cite{sirohi2021efficientlps} & 59.2 & 65.1 & 69.8 & 75.0 & 58.0 & 68.2 & 78.0 & \textbf{60.9} & {71.0} & 72.8 & 64.9 & -- & -- \\
    & GP-S3Net~\cite{razani2021gp} & \textbf{63.3} & \textbf{71.5} & \textbf{75.9} & \textbf{81.4} & \textbf{70.2} & \textbf{80.1} & {86.2} & {58.3} & \textbf{72.9} & \textbf{77.9} & \textbf{73.0} & -- & -- \\
    & \revision{Location Guided}~\cite{xian2022location} & 59.0 & 63.1 & 69.4 & 78.7 & 65.3 & 73.5 & \textbf{88.5} & 53.9 & 66.4 & 71.6 & 61.4 & -- & -- \\
    & \revision{CPGNet~\cite{li2022cpgnet}} & {62.7} & {67.5} & -- & -- & {70.0} & {--} & {--} & {57.3} & \textbf{--} & \textbf{--} & {67.4} & -- & -- \\
    & \revision{Panoptic-PHNet~\cite{li2022panoptic}} & {61.7} & -- & -- & -- & {69.3} & -- & -- & -- & -- & -- & {65.7} & -- & --\\
    & \revision{LCPS~\cite{zhang2023lidar}} & {59.0} & {68.8} & {68.9} & {79.8} & -- & -- & -- & -- & -- & -- & {63.2} & -- & --\\
    %
    %
    & 4DPLS~\cite{aygun21cvpr} & {56.5} & {61.9} & {66.8} & 79.0 & 57.3 & 64.3 & {88.0} & 56.0 & 68.6 & {72.4} & {64.8} & 63.3 & 75.3\\

    \midrule
    \midrule
    \parbox[t]{2mm}{\multirow{5}{*}{\rotatebox{90}{Ablations}}} & \hlps (class-specific) & {58.7} & {64.1} & {68.8} & 79.8 & 62.2 & 68.7 & 89.8 & 56.1 & 68.8 & {72.5} & {65.0} & 79.7 & {69.3}\\
    & \hlps (class-agnostic) & {58.7} & {64.1} & {68.8} & 79.8 & 62.3 & 68.8 & 89.8 & 56.1 & 68.8 & {72.5} & {65.0} & 79.9 & 69.3\\
    %
    & \hlps (+ holistic classifier) \quad\quad\quad & 58.8 & 64.2 & 68.9 & 79.8 & 62.5 & 68.9 & 89.8 & 56.1 & 68.8 & 72.5 & 65.0 & 80.0 & 69.6 \\
    & \textbf{\hlps (+ regression loss)} & 58.9 & 64.3 & 69.0 & 79.8 & 62.8 & 69.2 & 89.9 & 56.1 & 68.8 & 72.5 & 65.0 & 80.0 & 70.0 \\
    & \textbf{\hlps (+ majority vote)} & 59.0 & 64.4 & 68.9 & 80.0 & 63.0 & 69.1 & 90.3 & 56.1 & 68.8 & 72.5 & 64.2 & 81.9 & 67.4\\
    %

    \midrule
    \midrule
    %
    \parbox[t]{2mm}{\multirow{3}{*}{\rotatebox{90}{Oracle}}} & \hlps (obj. oracle) & {59.2} & {64.6} & {69.2} & 80.0 & 63.5 & 69.8 & 90.2 & 56.1 & 68.8 & 72.5 &  65.0 & 80.3 & 70.9\\
    & OWL (+ majority vote) & {59.7} & {65.1} & {69.4} & 80.6 & 64.6 & 70.2 & 91.8 & 56.1 & 68.8 & 72.5 & 64.2  & 82.3 & 69.1\\
    %
    %
    %
    %
    
    & \hlps (GT semantic map) & {98.3} & {98.3} & {99.2} & 99.0 & 96.1 & 98.3 & 97.8 & 100.0 & 100.0 & {100.0} & {100.0} & 99.4 & 97.2\\
    \bottomrule
    \end{tabular}
    }
    \label{tab:bench_semkitti}
\end{table*}

{\bf Semantic oracle.} With the semantic oracle experiment, we aim to answer the question: \textit{how far we can get in \lps with our baseline}? 
To answer this, we replace our learned classification network with ground-truth semantic maps (\textit{GT semantic map}), available in the validation set, but retain our instance branch. 
This yields a near-perfect PQ of $98.3\%$. As evident, we can recall $97.2\%$ of \thing objects with $99.4\%$ precision. 
This experiment raises questions of whether point-to-instance grouping needs to be learned; \textit{nearly all labeled instances are already included in our segmentation tree}. Moreover, this experiment suggests that future efforts should focus on driving further the \textit{point-classification} performance and that \textit{semantic labeling may be sufficient} -- near-perfect instance labels can be obtained via hierarchical clustering.

{\bf Objectness oracle.} 
Given a KPConv-based semantic network, \textit{what performance can we obtain with \textit{perfect} objectness scoring function?} 
To answer this, we use GT instance labels to score all segments as IoU between a segment and its best-matching GT segment (\textit{obj. oracle}). 
With this approach, we obtain $59.2\%$ PQ. By propagating semantic \textit{majority vote} within each segment, we can further improve precision and recall and, consequently, PQ ($+0.5\%$).  
This is an upper bound that we can obtain with the KPConv semantic backbone.

{\bf Ablations.} 
We start with a variant of our network with semantic and objectness heads (\cf, \cite{aygun21cvpr}). In this case, we compute per-segment objectness by averaging per-point objectness scores (see Sec.~\ref{sec:hlps}). 
The \textit{class-specific} variant builds the segmentation tree separately for each semantic class, while \textit{class-agnostic} variant builds it upon all \thing and \other points, effectively dropping fine-grained semantic information. With both, we obtain a PQ of $58.7\%$, $+2.2\%$ improvement over 4D-PLS, which uses identical segmentation and objectness networks for inference. 
In the class-agnostic variant, we observe improvement of $+0.2\%$ in terms of precision ($79.9\%$). We conclude this approach is more accurate because it is less sensitive to errors in semantic classification (\eg, a truck, part classified as truck and part as car, cannot be holistically segmented with \textit{class-specific} variant). That said, we can safely treat \thing and \other classes in a unified manner. 
Next, instead of averaging per-point objectness, we train a holistic second-stage objectness classifier using cross-entropy loss (see Sec.~\ref{sec:hlps}) and observe PQ improvement ($58.8\%$, $+0.1\%$). This way we gain an additional $+0.1\%$ in PQ. We hypothesize that by training our network using regression loss, we obtain smoother objectness scores compared to sharp-peaked (and overconfident) binary classification scores, which is beneficial for the tree-cut algorithm. 

\subsection{Open-World Lidar Panoptic Segmentation}
\label{sec:experiments:glps}
We now study \textit{within}-dataset performance on SemanticKITTI and \textit{cross}-dataset performance on SemanticKITTI $\to$ \pinkbgr{KITTI360} for two different source-domain vocabularies in Tab~\ref{tab:generalization}. \textit{Vocabulary 1} merges rarer classes into a catch-all \other class (as discussed in Sec.~\ref{sec:glps_eval}) while \textit{Vocabulary 2} closely follows the original SemanticKITTI class definitions. 
\revision{We provide further details on the construction of these vocabularies in Appendix~\ref{sec:sup:lipsow}.}
We report results of known \thing and \stuff classes using Panoptic Quality and mean-IoU, and for \unknown class (\revision{\ie points classified as \other during inference on} \pinkbgr{KITTI360}), we report Unknown Quality (UQ), Recall, and IoU. 
\revisiontwo{We analyze open-world generalization based on \textit{cross}-dataset performance (\ie \unknown) and not on \textit{within}-dataset performance.
}

\subsubsection{Baselines}
We compare our \hlps to vanilla 4D-PLS~\cite{aygun21cvpr}, trained in a single-scan setting. This network uses the same KPConv backbone~\cite{Thomas19ICCV}. 
We also train PolarSeg-Panoptic~\cite{zhou2021panoptic}, one of the top-performers on standard \lps (see Tab.~\ref{sec:lps}) for which \revision{\textit{source code is available}}. 
4D-PLS$^{\dagger}$ modifies the inference procedure: for points classified as \other, we lower the center-objectness threshold. We provide details in the appendix (Sec.~\ref{sec:sup:implementation}). This is based on the intuition that the objectness head of 4D-PLS should be able to generalize to novel classes, but with lower confidence. 
Finally, \hlps$^\ddagger$ is a modified variant of \hlps that uses a learned point-grouping mechanism (\cf, \cite{aygun21cvpr}) for \thing classes, and hierarchical segmentor for the \other class (\ie, does not treat instance segmentation of \thing and \other classes in a unified manner). This baseline is inspired by~\cite{wong2020identifying}. However, we use 4D-PLS as a backbone network and bottom-up grouping as described in Sec.~\ref{sec:hlps}.

\subsubsection{Lidar Semantic Segmentation}

We compare our approach to OSeg~\cite{cen2022open}, a state-of-the-art method for open-set Lidar Semantic Segmentation (LSS) in Tab.~\ref{tab:os-lss}. OSeg \textit{only} tackles the semantic point classification of Lidar scans and does not address the instance segmentation aspect of Lidar Panoptic Segmentation. OSeg consists of two stages: (i) Open-set semantic segmentation, where each point is classified into one of $K$ known or a catch-all class using redundant classifiers, and (ii) Incremental learning, where the catch-all categories are incorporated into the model. 
For a fair comparison with our setting, we report the performance of the first stage of OSeg, without performing incremental learning. 
We validate performance using our proposed vocabulary splits on SemanticKITTI using publicly available implementation (details in the appendix). As OSeg performs only point classification (\ie, semantic segmentation), we can compare to OSeg only in terms of mean intersection-over-union with our base network for point classification (4DPLS). 
Results show that OSeg \cite{cen2022open} significantly underperforms compared to our base 4DPLS network on both \known and \revision{\unknown}. OSeg performance on the \revision{\unknown} class performances may be caused by the object synthesis, which implicitly assumes the other category consists of only \thing and not \stuff classes, violating the spirit of LiPSOW. As can be seen, our 4DPLS-based network additionally outperforms OSeg on \known classes.

\begin{table}
    \centering
    \resizebox{1.0\linewidth}{!}{%
    \begin{tabular}{l|lcc}
    \toprule
    Vocabulary & Method & \known mIoU & \other IoU \\
    \hline
    Vocabulary 1 & 4DPLS~\cite{aygun21cvpr} & 79.8 & 56.9 \\
            & PolarSeg-Panoptic~\cite{zhou2021panoptic} & 74.2 & 47.3 \\
            & OSeg~\cite{cen2022open} & 67.3 & 1.5 \\
    \hline
    Vocabulary 2 & 4DPLS~\cite{aygun21cvpr} & 70.5 & 50.8 \\
            & PolarSeg-Panoptic~\cite{zhou2021panoptic} & 63.7 & 40.0 \\
            & OSeg~\cite{cen2022open} & 57.6 & 0.5 \\
    \bottomrule
    \end{tabular}
    }
    \caption{\textbf{Results of Lidar Semantic Segmentation}. Methods are trained on SemanticKITTI under the specified vocabulary, and evaluated on the SemanticKITTI validation set. Experiments indicate that OSeg~\cite{cen2022open}) struggles to generalize to held-out \other classes.
    }
    \label{tab:os-lss} 
\end{table}

\begin{table*}[ht]
    \centering
    \small
    \setlength{\tabcolsep}{3pt}
    \caption{\small \textbf{\gpkittilong}. 
    %
    The table reports \textit{within}-dataset performance (SemanticKITTI), and \textit{cross}-dataset performance (\pinkbgr{KITTI360}) for two class ontologies, where \textit{Vocabulary 1} merges rarer classes into a catch-all \other class while \textit{Vocabulary 2} closely follows original SemanticKITTI class definitions.
    %
    %
    Interestingly, OWL outperforms baselines even for \known classes ($69.4\%$ PQ compared with $67.8\%$ 4D-PLS in-domain and $59.4\%$ vs. $56.1\%$ PQ cross-domain for \textit{Vocab. 1}). For \revision{new \unknown classes at test-time}, \hlps recalls $45.1\%$ of labeled instances ($36.3\%$ UQ) while a modified version of 4D-PLS can only recall $6.0\%$, leading to $4.2\%$ UQ. We observe similar trends with \textit{Vocab. 2}. 
    %
    Note that \textit{Vocab. 1} generalizes much better across datasets, suggesting that grouping rare classes in a catch-all {\tt other} during training leads to better generalization. 
    These results also suggest that we do not need a separate geometric clustering module for \other classes, as suggested in~\cite{wong2020identifying}. Such an approach is similar to \hlps$^\ddagger$ (which uses a learned instance grouping for \thing classes and a separate hierarchical segmentation tree for \other), which does not improve performance. %
    %
    }
    \resizebox{1.0\linewidth}{!}{%
\begin{tabular}{ccl|ccc|cc|c|cc||ccc|c}
\toprule
\centering
& & \multirow{2}{*}{Method} & \multicolumn{8}{c}{\known} & \multicolumn{4}{c}{\unknown}\tabularnewline
& & & PQ & RQ & SQ & PQ$^{Th}$ & PQ$^{St}$ & mIoU & Precision$^{Th}$ & Recall$^{Th}$ & UQ & Recall & SQ & IoU \tabularnewline
%
%
\midrule\parbox[t]{2mm}{\multirow{10}{*}{\rotatebox{90}{Vocabulary 1}}} & 
 \parbox[t]{2mm}{\multirow{5}{*}{\rotatebox{90}{SemKITTI}}} & 4D-PLS~\cite{aygun21cvpr} & 67.8 & 78.5 & 85.4 & 60.0 & 71.7 & 79.8 & 66.1 & 85.4 & 7.8 & 10.8 & 71.9 & 56.9 \\
& & 4D-PLS$^{\dagger}$ & 67.6 & 78.3 & 85.4 & 59.4 & 71.7 & 79.8 & 65.5 & 85.1 & 19.9 & 27.6 & 72.0 & 56.9 \\
& & PolarSeg-Panoptic~\cite{zhou2021panoptic} \quad\quad\quad& 68.6  & \textbf{80.2} & 83.6 & \textbf{68.2} & 68.8 & 74.2 & \textbf{79.5} & 76.8 & 10.2 &14.7  & 69.3 &  47.3
\tabularnewline
& & OWL$^\ddagger$ & 67.8 & 78.5 & 85.4 & 60.0 & 71.7 & 79.8 & 66.1 & 85.4 & \textbf{39.8} & \textbf{48.5} & \textbf{82.1} & 56.9 \tabularnewline
& & OWL (Ours) & \textbf{69.4} & 79.5 & \textbf{86.3} & 64.7 & \textbf{71.7} & \textbf{79.8} & 69.3 & \textbf{87.6} & 39.6 & 48.4 & 81.8 & \textbf{56.9}\tabularnewline
\cmidrule{2-15}

& \parbox[t]{2mm}{\multirow{5}{*}{\rotatebox{90}{KITTI360}}} & \markc 4D-PLS~\cite{aygun21cvpr} & \markc 56.1 & \markc 67.4 & \markc 80.5 & \markc 56.2 & \markc 56.0 & \markc 65.3 & \markc 63.5 & \markc 70.8 & \markc 1.3 & \markc 2.0 & \markc 65.7 & \markc 11.4 \\
& & \markc 4D-PLS$^{\dagger}$ & \markc 55.8 & \markc 67.2 & \markc 80.4 & \markc 55.4 & \markc 56.0 & \markc 65.3 & \markc 62.5 & \markc 70.5 & \markc 4.2 & \markc 6.0 & \markc 70.6 & \markc 11.4 \\
& & \markc PolarSeg-Panoptic~\cite{zhou2021panoptic} & \markc 0.7 & \markc 0.9 & \markc 73.0 & \markc 0.7 & \markc 0.7 & \markc 1.6 & \markc 14.6 & \markc 0.6 & \markc 0.0 & \markc 0.1 & \markc 76.4 & \markc 9.0 \\
& & \markc OWL$^\ddagger$ &\markc 56.1 & \markc 67.4 & \markc 80.5 & \markc 56.2 & \markc 56.0 & \markc 65.3 & \markc 63.5 & \markc 70.8 & \markc \textbf{36.6} & \markc \textbf{45.4} & \markc \textbf{80.6} & \markc 11.4\tabularnewline
& & \markc OWL (Ours) & \markc \textbf{59.4} & \markc \textbf{70.3} & \markc \textbf{81.8} & \markc \textbf{66.2} & \markc \textbf{56.0} & \markc \textbf{65.3} & \markc \textbf{72.5} & \markc \textbf{78.5} & \markc 36.3 & \markc 45.1 & \markc 80.5 & \markc \textbf{11.4} \tabularnewline
\midrule
%
%
\midrule\parbox[t]{2mm}{\multirow{10}{*}{\rotatebox{90}{Vocabulary 2}}} & \parbox[t]{2mm}{\multirow{5}{*}{\rotatebox{90}{SemKITTI}}} & 4D-PLS~\cite{aygun21cvpr} & 60.2 & 69.3 & 81.4 & 57.9 & 61.4 & 70.5 & 70.2 & 72.9 & 16.4 & 22.2 & 73.8 & 50.8 \\
& & 4D-PLS$^{\dagger}$ & 60.1 & 72.7 & 81.4 & 57.6 & 61.4 & 70.5 & 69.9 & 72.8 & 20.7 & 29.0 & 71.3 & 50.8 \\
& & PolarSeg-Panoptic~\cite{zhou2021panoptic} & 58.6 & 68.2 & 79.4 & 56.5 & 55.2 & 63.7 & 72.9 & 63.7 & 14.9 & 20.7 & 72.1 & 40.0 \tabularnewline
& & OWL$^\ddagger$ & 60.2 & 69.9 & 81.4 & 57.9 & 61.4 & 70.5 & 70.2 & 72.9 & 49.2 & 56.9 & \textbf{86.5} & 50.8 \tabularnewline
& & OWL  (Ours)& \textbf{61.9} & \textbf{73.9} & \textbf{82.3} & \textbf{62.9} & \textbf{61.4} & \textbf{70.5} & \textbf{74.3} & \textbf{75.2} & \textbf{49.3} & \textbf{57.0} & 86.3 & \textbf{50.8} \tabularnewline
\cmidrule{2-15}
%
& \parbox[t]{2mm}{\multirow{5}{*}{\rotatebox{90}{KITTI360}}} & \markc  4D-PLS~\cite{aygun21cvpr} & \markc 45.9 & \markc 56.8 & \markc 77.5 & \markc 47.9 & \markc 44.8 & \markc 53.4 & \markc 57.3 & \markc 59.1 & \markc 3.4 & \markc 4.9 & \markc 69.8 & \markc 6.3 \\

& & \markc 4D-PLS$^{\dagger}$ & \markc 45.8 & \markc 56.7 & \markc 77.5 & \markc 47.7 & \markc 44.8 & \markc 53.4 & \markc 56.9 & \markc 59.1 & \markc 5.5 & \markc 7.9 & \markc 69.6 & \markc 6.3  \\

& & \markc PolarSeg-Panoptic~\cite{zhou2021panoptic} & \markc 1.2 & \markc 1.8 & \markc 61.7 & \markc 0.4 & \markc 1.6 & \markc 3.0 & \markc 9.2 & \markc 0.3 & \markc 0.0 & \markc 0.0 & \markc 71.1 & \markc 0.7 \\

& & \markc OWL$^\ddagger$ & \markc 45.9 & \markc 56.8 & \markc 77.5 & \markc 47.9 & \markc 44.8 & \markc 53.4 & \markc 57.3 & \markc 59.1  & \markc \textbf{21.3} & \markc \textbf{26.9} & \markc 79.0 & \markc 6.3 \\

& & \markc OWL  (Ours) & \markc \textbf{53.4} & \markc \textbf{64.2} & \markc \textbf{79.9} & \markc \textbf{55.1} & \markc \textbf{52.5} & \markc \textbf{53.4} & \markc \textbf{64.4} & \markc \textbf{64.7} & \markc 21.2 & \markc 26.8 & \markc \textbf{79.0} & \markc \textbf{6.3} \\
\bottomrule
\end{tabular}
}
    \label{tab:generalization}
\end{table*}

\subsubsection{Lidar Panoptic Segmentation}

\PAR{Vocabulary 1.} In Tab.~\ref{tab:generalization}, \hlps is top-performer for \known classes ($+1.6\%$ \wrt 4D-PLS and $+0.8\%$ \wrt PolarSeg-Panoptic). 
Similarly for \other, \hlps recalls $48.4\%$ of objects, compared to $10.8\%$ recalled by 4D-PLS and $14.7\%$ by PolarSeg-Panoptic. 
Note that the only \other objects with instance labels in SemanticKITTI are {\tt bicycle} and {\tt motorcycle}. 
For the \revision{\unknown} class in the cross-domain section (\pinkbgr{KITTI360}), we observe 4D-PLS recalls only $2.0 \%$ of instances ($1.3\%$ UQ). By changing the inference, 4D-PLS$^\dagger$ recalls $6.0\%$ of objects. 
\hlps recalls $45.1\%$ of objects, leading to $36.3\%$ UQ. 
This suggests there is a significant potential to improve further without modifying our instance segmentor: the bottleneck seems to be the point-level classifier (see low IoU of $11.4\%$). 
This result also highlights that SemanticKITTI by itself is not sufficient for studying open-world \lps methods due to a limited number of classes with instance labels. In \pinkbgr{KITTI360} we have a significantly larger number of instances in the \revision{\unknown} class, exposing the poor generalization of methods trained to segment only the $K$-known classes.  

In SemanticKITTI$\to$\pinkbgr{KITTI360}, we observe a performance drop in cross-domain evaluation, including \known classes. 4D-PLS performance drops from $67.8\%$ (SemanticKITTI) to $56.1\%$ PQ (KITTI360) and $79.8\%\to65.3\%$ mIoU. PolarSeg-Panoptic works well when evaluated in a within-domain setting but fails to generalize to KITTI360 ($0.7\%$ PQ). This suggests that existing models are very sensitive to data distribution shifts. Future efforts should aim not only to improve point classification performance within-domain but also in cross-domain settings. 

\begin{table*}[ht]
    \centering
    \small 
    \setlength{\tabcolsep}{3pt}
    \caption{\small \textbf{Per-class results (\textit{Vocabulary 2})} \wrt PQ for known classes and UQ for the {\tt unknown}. We observe performance drop for rarer classes (\eg, {\tt motorcycle}) and among \stuff  (\eg, {\tt building}, {\tt vegetation}) that may look differently in different areas of the city.
    }
   \resizebox{1.0\linewidth}{!}{%
\begin{tabular}{clccccccccccccccc|c}
\toprule
&    & \thc{\texttt Car} & \thc{\texttt Hum.} & \thc{\texttt Truck} & \thc{\texttt Moto.} & \thc{\texttt Bic.} & \thc{\texttt Traf. Sign}  & \thc{\texttt Trunk} & \thc{\texttt Pole} & \stc{\texttt Rd.} & \stc{\texttt Build.} & \stc{\texttt Veg.} & \stc{\texttt Fen.} & \stc{\texttt Sidew.} & \stc{\texttt Ter.} & \stc{\texttt Park.}  & \unknown \tabularnewline
\midrule 
 
\parbox[t]{2mm}{\multirow{3}{*}{\rotatebox{90}{SemK.}}} 
& 4D-PLS~\cite{aygun21cvpr}  & 82.2 & 68.0 & 44.1 & 59.4 & 35.9 & \unpub{54.4} & \unpub{60.7} & \unpub{60.7} & 94.0 & 86.6 & 86.7 & 22.0 & 77.7 & 58.4 & 21.9 & 16.4 \tabularnewline
& PolarSeg-Panoptic~\cite{zhou2021panoptic}  & 85.4 & 52.6 & 52.5 & 52.6 & 39.5 & \unpub{47.5} & \unpub{33.4} & \unpub{49.0} & 90.7 & 84.8 & 85.4 & 16.4 & 70.6 & 53.3 & 20.7 & 14.9 \tabularnewline
& \hlps  & 90.6 & 68.0 & 50.2 & 65.4 & 40.3 & \unpub{54.4} & \unpub{51.7} & \unpub{60.7} & 94.0 & 86.6 & 86.7 & 22.0 & 77.7 & 58.4 & 21.9 & 49.3\tabularnewline
%
%
\midrule
\midrule
\parbox[t]{2mm}{\multirow{3}{*}{\rotatebox{90}{K.360}}} 
& \markc 4D-PLS~\cite{aygun21cvpr}  & \markc 81.5 &\markc 52.4 &\markc 41.9 & \markc 43.4 &\markc  20.3 &\markc  47.0 & \markc  NA & \markc 20.9 & \markc 89.8 & \markc 60.6 & \markc 77.2 &\markc 9.6 & \markc 63.0 & \markc 23.5 & \markc 11.3 & \markc 3.4\tabularnewline
& \markc PolarSeg-Panoptic~\cite{zhou2021panoptic}  &\markc 1.7  &\markc 0.0 & \markc 0.4&\markc 0.0 & \markc 0.0 & \markc 2.7 & \markc NA & \markc 1.0 & \markc 0.6 & \markc 0.8 & \markc 1.0 &\markc 0.0 &\markc 0.2 & \markc 0.1 & \markc 0.0 & \markc 0.0\tabularnewline
& \markc \hlps & \markc 88.3 & \markc 62.1 & \markc 55.8 & \markc 45.8 & \markc 23.2 &\markc 74.3 & \markc NA & \markc 63.2 & \markc 89.8 & \markc 60.6 & \markc 77.2 & \markc 9.6 & \markc 63.0 & \markc 23.5 & \markc 11.3 &\markc 21.2\tabularnewline
\bottomrule
\end{tabular}
    }
    \label{tab:bench_per_class_TS2}
\end{table*}

\PAR{Vocabulary 2.} 
This setting follows more closely the official SemanticKITTI vocabulary and exposes a smaller number of semantic classes as instances of the \other class during training. We observe that \textit{Vocabulary 1} generalizes much better across datasets, suggesting that grouping rare classes in a catch-all \other class leads to better generalization. 

\begin{figure*}[t]
    \centering
    \includegraphics[width=0.99\linewidth]{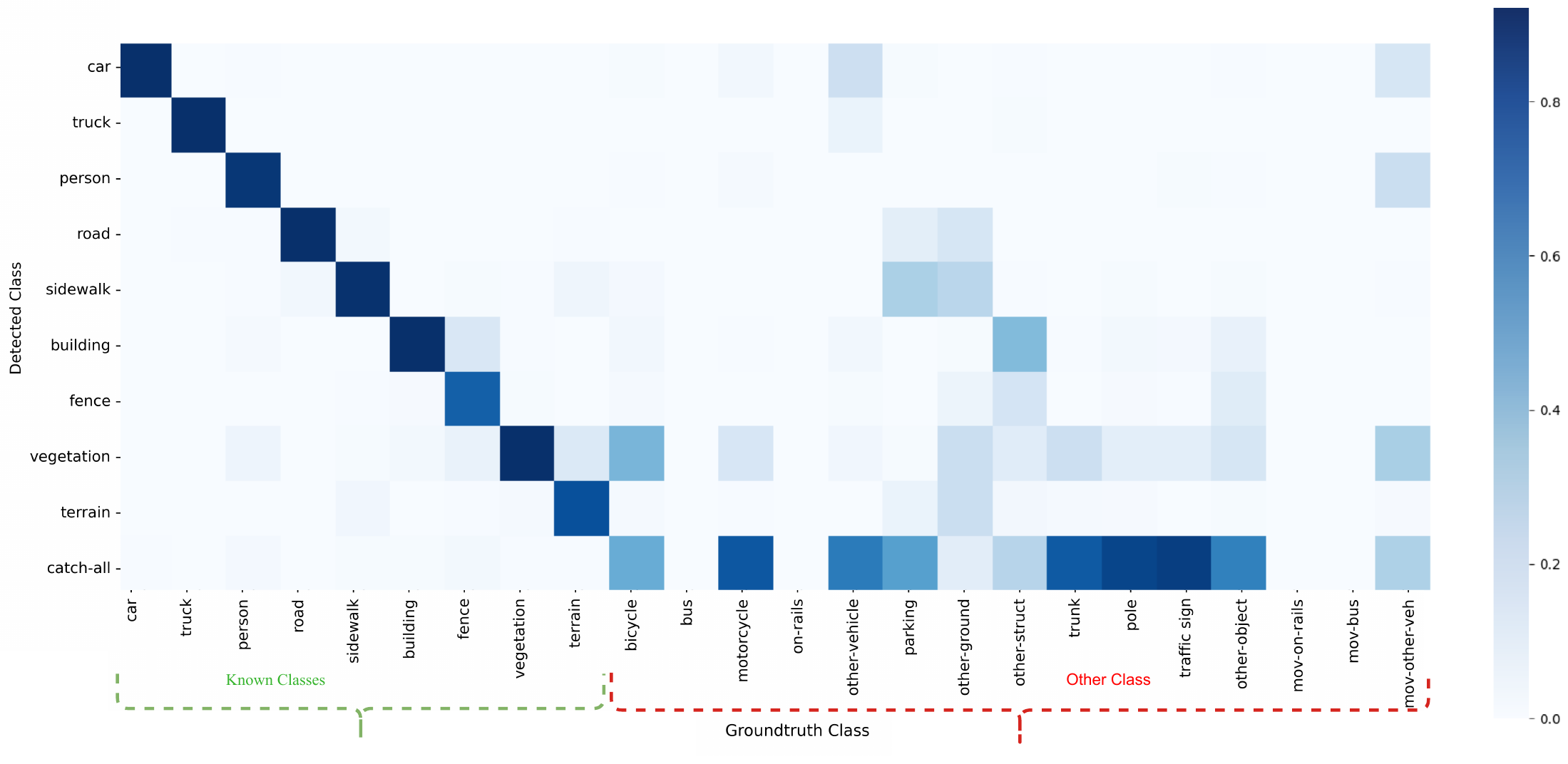}
    \caption{The extended confusion matrix for \hlps trained on SemanticKITTI and evaluated in-domain (on SemanticKITTI), using \textit{Vocabulary 1}. On the left side, we see the confusion among \known classes. On the right, we can see which \known classes are confused with classes that form the \other class. For \known classes, we observe a confusion between (related) terrain and vegetation. We also observe that several \other points are misclassified as \known.
    Class \texttt{other-vehicle} is often misclassified as \texttt{car} or \texttt{truck}, while \texttt{other-ground} and \texttt{parking} are commonly misclassified as \texttt{sidewalk} and \texttt{road} classes. This explains the low IoU, observed in Table 2 (main paper) on \other in SemanticKITTI.
    }
    \label{fig:conf_mat_semkitti}
\end{figure*}

\begin{figure*}[t]
    \centering
    \includegraphics[width=0.99\linewidth]{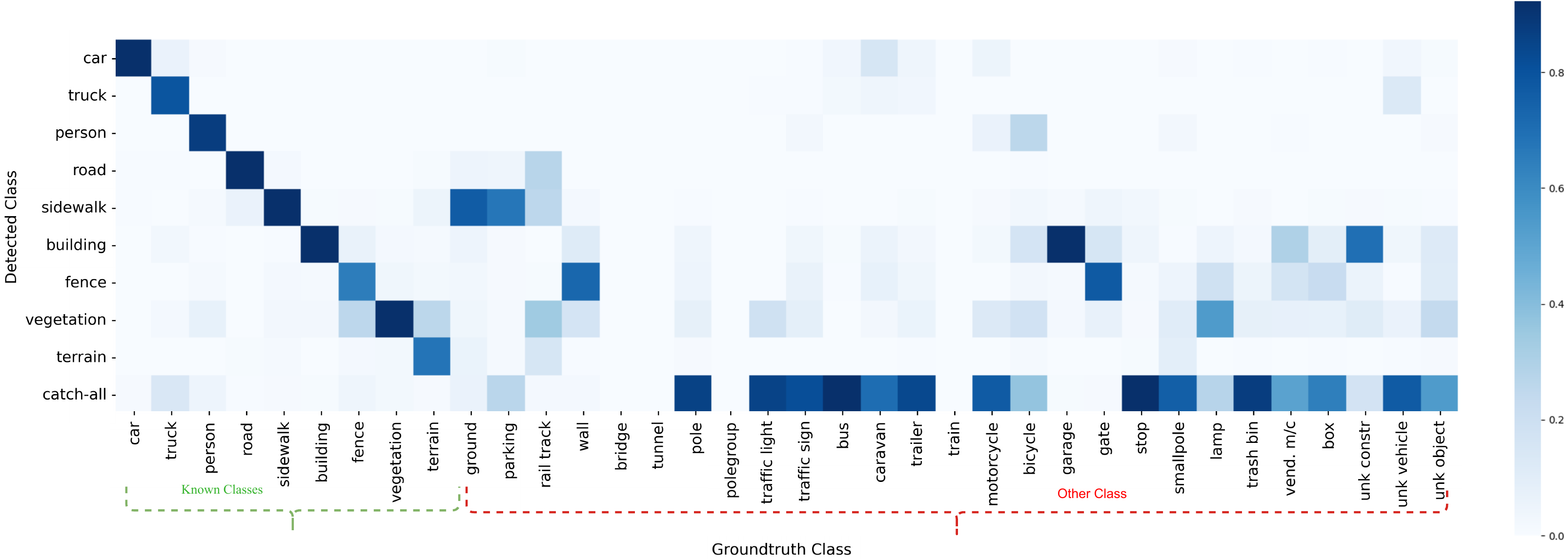}
    \caption{The extended confusion matrix for \hlps trained on SemanticKITTI and evaluated in cross-domain setting (on KITTI-360), using \textit{Vocabulary 1}. On the left side, we see the confusion among \known classes. On the right, we can see which \known classes are confused with classes that form the \other class. 
    Contrary to the in-domain confusion, we observe more confusion within \known classes. For instance, \texttt{car} and \texttt{truck} classes are often confused. The class \texttt{sidewalk} is often misclassified as \texttt{terrain}, while almost all \known classes are confused with \texttt{vegetation}.
    As can be seen, there is confusion between \known and \unknown classes.
    \texttt{Ground} and \texttt{parking} are often predicted as \texttt{road} and \texttt{sidewalk}. Class \texttt{wall} (a novel \other-\stuff class) is confused with \texttt{fence}, building, and vegetation, presumably due to their geometric similarity. Class \texttt{trailer} is frequently confused with class \texttt{car}. 
    As demonstrated, cross-domain semantic segmentation is a challenging problem.}
    \label{fig:conf_mat_kitti360}
\end{figure*}

\PAR{SemanticKITTI$\to$\pinkbgr{KITTI360} gap.} Why does performance drop from SemanticKITTI to KITTI360, even though both were recorded in the same city, with the same sensor? To answer this question, we analyze per-class performance in Tab.~\ref{tab:bench_per_class_TS2} (\textit{Vocabulary 2}) and confusion tables (Fig~\ref{fig:conf_mat_kitti360}). 
For common \thing classes, there is a minimal performance drop ($90.6\to88.3$ PQ for \texttt{car}). Indeed \textit{cars should look identical in different regions of the city}. As expected, we observe a performance drop for rarer \thing classes ($65.4\to 45.8$ for \texttt{motorcycle}, $40.3 \to 23.2$ for \texttt{bicycle}), as only a handful of instances of these classes are observed in the (dominantly static) SemanticKITTI. 
We observe larger performance drops for \stuff classes (\eg, \texttt{building}, \texttt{vegetation}, \texttt{fence}, \texttt{terrain}). The reason for this drop is two-fold. Firstly, KITTI360 covers a larger area of the city that does not overlap with SemanticKITTI. Therefore, in KITTI360, we observe a larger diversity of these classes, which confuses the semantic classifier. Second, these classes are often confused with \stuff classes, labeled only in KITTI360. For example, \texttt{building} is commonly confused with KITTI360 classes \texttt{garage} and \texttt{wall}, and \texttt{fence} is confused with \texttt{wall} and \texttt{gate}. Class \texttt{thrash bin} is often confused with a \stuff class \texttt{sidewalk}, likely due to context: thrash bins are usually seen on sidewalks. Outlier synthesis~\cite{cen2022open,kong2021opengan} could be used to minimize this confusion in the future.


\subsection{Confusion Analysis}
\label{sec:confusion}

To further analyze the per-class semantic segmentation performance, we plot extended confusion matrices, similar to those reported in open-set object detection~\cite{dhamija20WACV}. The horizontal axis represents the ground-truth classes, and the vertical axis represents \hlps predictions. We extend the \other class into its fine-grained split on the horizontal axis. 
Since \textit{Vocabulary 1} consists of more held-out classes, we visualize the matrices for this setting. 

\PAR{SemanticKITTI.} In Fig.~\ref{fig:conf_mat_semkitti}, we show the confusion of \hlps on SemanticKITTI. Among \known classes, we observe that the \texttt{terrain} class is most confused with the \texttt{vegetation} class. In the \other class, we observe significant confusion with the \known classes. For example, \texttt{other-vehicle} is often misclassified as \texttt{car} or \texttt{truck}. Furthermore, \texttt{other-ground} and \texttt{parking} are often misclassified as \texttt{sidewalk} and \texttt{road}. Confusion most commonly arises between classes with super-classes.

\PAR{KITTI360.} In Fig.~\ref{fig:conf_mat_kitti360}, we visualize the extended confusion matrix for \hlps on KITTI360. In the \known classes, we observe confusion between the \texttt{car} and \texttt{truck} classes. Furthermore, \texttt{fence} and \texttt{terrain} are also frequently misclassified as \texttt{vegetation}.
Next, we analyze confusion between \known and \other classes. We observe that \texttt{ground} and \texttt{parking} are often misclassified as \texttt{road} and \texttt{sidewalk}. \texttt{Wall}s are confused with the \texttt{fence}, \texttt{building}, and \texttt{vegetation}, while \texttt{trailer} is commonly confused with \texttt{car}. This demonstrates the challenge of open-world generalization of semantic segmentation, indicating the need for
future research on this front.

\begin{figure*}
\includegraphics[width=0.99\linewidth]{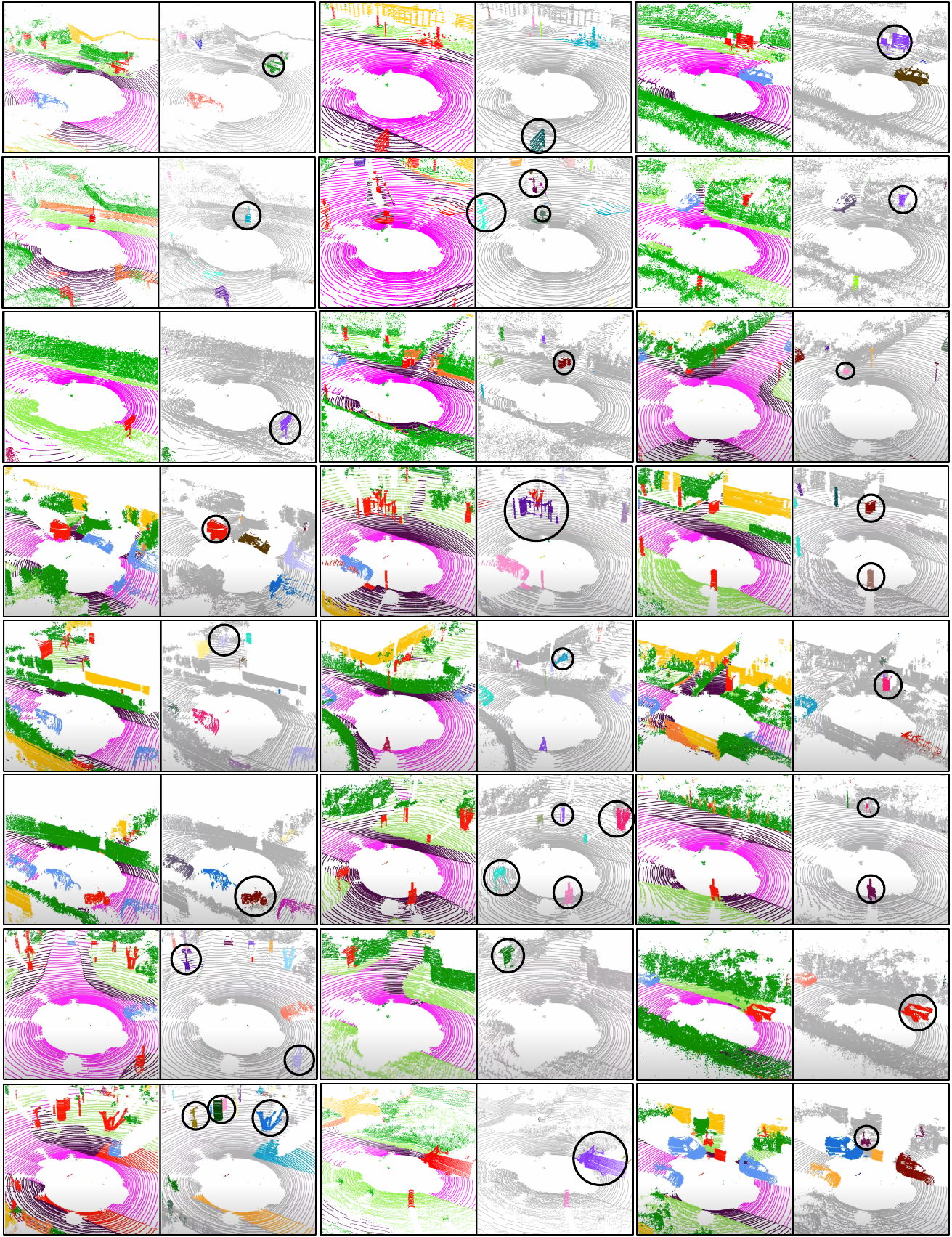}
    \caption{\textbf{Qualitative results on KITTI360 dataset.} \hlps successfully segments several \other objects (\textit{left}, shown in {\color{red} red}; \textit{right}: segmented instances). \\
    Issues and challenges: we observe \hlps occasionally under-segments \other instances (see, \eg, \textit{top} row).}
    \label{fig:owl_vis}
\end{figure*}

\subsection{Qualitative Results}
\label{sec:qualitative}

\revision{
In Fig.~\ref{fig:owl_vis}, we show several visual examples for \hlps in the KITTI360 dataset (\textit{Vocabulary 1}), focusing on instances of \revision{\unknown} classes. As can be seen, \hlps performs well in challenging cases with several \revision{\unknown} objects in the same scene. For example, in \textit{sixth-row-middle} and \textit{eighth-row-left}, our method segments common objects, such as \texttt{trunk}, \texttt{sign board}, and \texttt{pole}. Moreover, \hlps segments several rarer objects. For example, in \textit{first-row-left}, \hlps segments a \texttt{wheelbarrow}. In \textit{first-row-right}, \hlps segments a \texttt{bus stop}. In \textit{fifth-row-left}, \textit{fifth-row-middle} and \textit{sixth-row-left}, \hlps segments a \texttt{swing}, \texttt{stoller}, and a \texttt{motorcycle}.
}

\revision{
However, we also observe failure cases. In the top row of Fig~\ref{fig:owl_vis}, \hlps sometimes under-segment nearby objects. For example, on \textit{first-row-mid} and,\textit{first-row-right} neighboring poles and signs are clustered as a single instance.
}

\revision{
To further showcase the performance of our method, we provide a \href{https://drive.google.com/file/d/1sAXrzIeEXbowAvMuqXDLyuugyGhUnfM5/view?usp=sharing}{video} with \hlps inference on SemanticKITTI and KITTI360 lidar sweeps.
}

\section{Discussion and Conclusion}

We investigate Lidar Panoptic Segmentation in an Open World setting (LiPSOW), for which we set up baselines and an evaluation protocol. 
We demonstrate that our \hlps performs significantly better than prior work for in-domain and cross-domain evaluations. In addition to better generalization across domains, \hlps segments a large number of instances in the \other class.
Finally, we observed that grouping rare classes into a catch-all {\tt other} class leads to significantly better cross-domain generalization. We hope our insights spark future investigation and help build perception models that can generalize to novel environments.



We envision LiPSOW as a first stage towards an end-to-end continual learning paradigm where unknown objects from lidar scans are discovered online and clustered offline. Based on discovered object clusters (with human-in-the-loop annotations to provide further categorical refinement), the network can be updated using incremental/continual learning. Such signals can be incorporated into trajectory prediction or motion planning algorithms to enable safe maneuvers.


\backmatter

\bmhead{Supplementary information}

We provide further discussion about our method, evaluation, and \revision{complexity analysis} in the appendix. 







\section*{Declarations}


\begin{itemize}
\item Funding: This work was supported by the CMU Argo AI Center for Autonomous Vehicle Research.
\item Competing interests: The authors have no competing interests to declare that are relevant to the content of this article.
\item Ethics approval: N/A
\item Consent to participate: N/A
\item Consent for publication: N/A
\item Availability of data and materials: All experiments make use of publicly available datasets (SemanticKITTI, Kitti360). 4DPLS~\cite{aygun21cvpr}, PolarSeg-Panoptic~\cite{zhou2021panoptic}, and OSeg~\cite{cen2022open} experiments use open-source code.
\item Code availability: Code has been released along with the submission.
\item Authors' contributions: Experiments were performed by Anirudh Chakravarthy and Meghana Ganesina, who were advised by Aljosa Osep, Deva Ramanan, Shu Kong, and Laura Leal-Taixé. Peiyun Hu provided support with instance segmentation implementation. All authors approved the manuscript submission.
\end{itemize}


\clearpage
\begin{appendices}

\section{LiDAR Panoptic Segmentation in Open-World}
\label{sec:sup:lipsow}

\PAR{Vocabulary splits.} 
We detail our vocabulary split for SemanticKITTI~\cite{Behley19ICCV,Behley21icra} and KITTI360~\cite{Liao2021ARXIV} in Table~\ref{tab:sup:task_split}. 
The categorization of \stuff and \thing follows from the KITTI360 ontology.
\textit{Vocabulary 1} is constructed by sorting SemanticKITTI superclasses by the number of instances in a superclass and holding-out tail classes as \other. 
\textit{Vocabulary 2} is constructed by holding out only \textit{the rarest} object instances that do not correspond to any semantic class, labeled in SemanticKITTI (\eg, \texttt{other-vehicle}, \texttt{other-object}, \etc)

\PAR{Vocabulary consistency.} Inconsistent labeling policies across datasets cause label shifts. For this reason, we base our evaluation set-up on SemanticKITTI and KITTI360. The two datasets adopt largely consistent labeling policies and the same sensor; therefore, the shift in data distribution can be thought of as a result of new classes emerging across datasets. To ensure we have a consistent class vocabulary, we perform the following measures: 
\begin{itemize}
    \item We merge \texttt{rider} and \texttt{bicyclist} (SemanticKITTI) with \texttt{human} and \texttt{rider} (KITTI360) into a single \texttt{human} class to ensure consistency. 
    \item Classes \texttt{pole} and \texttt{traffic sign} are commonly treated as \thing classes. However, in SemanticKITTI, they are treated as \stuff classes because we do not have instance-level annotations for them. As instance labels for these classes are available in KITTI360, we treat them as \other \thing classes in KITTI360. Therefore, individual instances of these classes must be segmented. This is consistent with the overall goal of LiPSOW: methods must segment all instances, including those not labeled in SemanticKITTI. 
    \item We treat \texttt{building} as a \stuff class in SemanticKITTI and KITTI360 (\ie, we do not treat the building class as an object, a \thing class).  
\end{itemize}

\begin{table*}[t]
    \centering
    \small 
    \setlength{\tabcolsep}{3pt}
    \caption{\textbf{LiPSOW class ontology}. In our experimental section, we report results using two vocabularies: \textit{Vocabulary 1} \& \textit{Vocabulary 2}, both derived from the SemanticKITTI~\cite{Behley19ICCV,Behley21icra} class ontology. We color \stuff classes in \first{red} and \thing classes in \third{blue}. With \unpub{grey} we denote classes with instance labels available in KITTI360, but not in SemanticKITTI. LiPSOW methods have access to all labels from \known classes; however, no access to labels for \other.} 
    \begin{tabular}{l p{0.25\linewidth}|p{0.25\linewidth}|p{0.45\linewidth}}
    \toprule
     & \texttt{Known} & \texttt{Other} (SemanticKITTI) & \texttt{Unknown} (KITTI360)\\
    \midrule
    \parbox[t]{2mm}{\multirow{1}{*}{\rotatebox{90}{Vocab. 1}}} & \third {car, truck, human} \newline
    \first{road, sidewalk, fence, vegetation, terrain, building} & \third{bicycle, motorcycle, other-vehicle}, \unpub{trunk, pole, traffic-sign, other-structure, other-object} \newline 
    \first{other-ground, parking} &  \third {pole, pole group, traffic light, traffic sign, bus, caravan, trailer, train, motorcycle, bicycle, garage, stop, small pole, lamp, trash bin, vending machine, box, unk. construction, unk. vehicle, unk. object} \newline 
    \first{ground, parking, rail-track, wall, bridge, tunnel, gate} \\
    \midrule
    \midrule
    \parbox[t]{2mm}{\multirow{1}{*}{\rotatebox{90}{Vocab. 2}}} & \third{car, bicycle, motorcycle, truck, human},  \unpub{trunk, pole, traffic-sign} \newline  \first{road, sidewalk, fence, vegetation, terrain, parking, building}  & \third{other-vehicle}, \unpub{other-structure, other-object} \newline \first{other-ground} & \third{traffic light, bus, caravan, trailer, train, garage, stop, lamp, trash bin, vending machine, box, unk. construction, unk. vehicle, unk. object} \newline \first{ground, rail-track, wall, bridge, tunnel, gate} \\ 
    
    \bottomrule
    \end{tabular}
    \label{tab:sup:task_split}
\end{table*}

\PAR{Model training.}
We train all LiPSOW methods using instance labels for {\tt known-\third{things}} and semantic labels for \known-{\tt \third{things}} and {\tt known-\first{stuff}}. The instance labels for the \other classes are held out, \ie, not available during training.
The LiPSOW methods should classify these points as \other instead of performing a fine-grained semantic classification of these points. Additionally, LiPSOW methods need to segment novel instances from \other, \eg, segment \third{bicycles} in SemanticKITTI (\textit{Vocabulary 1}) and \third{vending machines} in KITTI360 (\textit{Vocabulary 1} \& \textit{Vocabulary 2}).

\begin{figure*}
    \centering
    \includegraphics[width=0.99\linewidth]{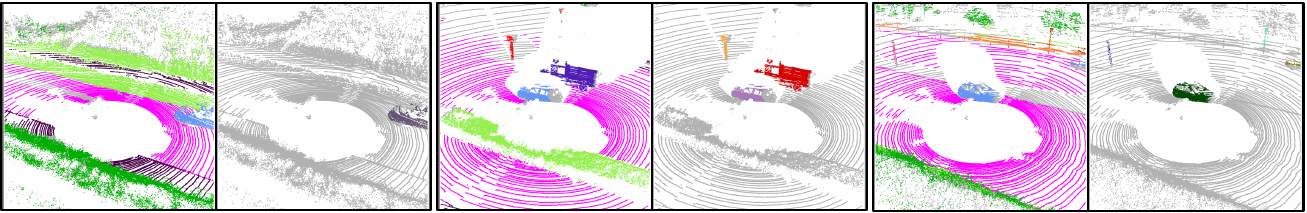}
    \caption{Semantic and instance labels for KITTI360 evaluation, retrieved from the dense accumulated labeled point clouds. Grey color denotes points for which we could not retrieve labeled points within a 10 cm radius to transfer labels.}
    \label{fig:sup:kitti360_labels}
\end{figure*}

\PAR{KITTI360 Ground-Truth.} To evaluate methods on the KITTI360 dataset, we require per-scan semantic and instance labels for Velodyne Lidar scans. However, KITTI360 only provides multiple accumulated point clouds (accumulated over approximately 200m), recorded by the SICK Lidar sensor. We use these dense accumulated point clouds to retrieve per-scan labels for individual Velodyne point clouds. Concretely, we use publicly available scripts~\cite{k360_label} to align individual Velodyne scans with the accumulated point cloud based on known vehicle odometry. Once aligned, we perform a nearest neighbor search for each Velodyne point in the corresponding accumulated point cloud. In case a match is not found within a 10cm radius, we mark this point as unlabeled (ignored during the evaluation). We visualize the retrieved labels for Velodyne point clouds in Fig~\ref{fig:sup:kitti360_labels}.

\section{Implementation Details}
\label{sec:sup:implementation}

\subsection{4DPLS$^\dagger$} 
Our method and several baselines are based on 4D-PLS~\cite{aygun21cvpr} that employs an encoder-decoder point-based KPConv~\cite{Thomas19ICCV} backbone for point classification and instance segmentation. The instance segmentation branch consists of three network heads. The \textit{objectness} head predicts for all points how likely they are to represent a (modal) instance center. The \textit{embedding} and \textit{variance} heads are used to associate points with their respective instance centers. During the inference, we select the point $p_i$ with the highest objectness, evaluate all points under a Gaussian (parameterized by the predicted mean and variances for $p_i$), and assign points to this cluster if the point-to-center association probability is higher than a threshold, \ie, $> 0.5$. This process is repeated until the maximum objectness is below a certain threshold ($0.1$ in 4D-PLS). To ensure high-quality segments, 4D-PLS also enforces that the highest objectness should be $> 0.7$.

Since the objectness head is trained independently of the semantic head in a class-agnostic fashion, we hypothesize it should be able to learn a general notion of objectness from geometric cues. Therefore, we evaluate whether it can segment instances of novel classes with lower confidence. 
Therefore, we adapt the inference procedure in 4D-PLS to allow additional instances to be segmented. We achieve this by reducing the minimum objectness threshold from $0.7$ to $0.3$ for \other, while maintaining the same threshold for {\tt known things}. Our experimental evaluation confirms that this baseline can segment a larger number of \other instances compared to 4D-PLS. 

\subsection{OSeg~\cite{cen2022open}} 

OSeg~\cite{cen2022open} introduces a novel strategy for open-world semantic segmentation of LiDAR Point Clouds. The proposed framework consists of two stages: (i) Open-Set semantic segmentation (OSeg) and (ii) Incremental Learning. For a fair comparison, we benchmark OSeg against our baselines.

OSeg introduces redundancy classifiers on top of a closed-set model to output scores for the \texttt{unknown class}. In addition, OSeg uses unknown object synthesis to generate pseudo-unknown objects based on real novel objects. The OSeg formulation considers \texttt{other vehicle} as a novel category for SemanticKITTI. To benchmark under our proposed LiPSOW formulation, we modify OSeg to allow for more classes in \other (based on our vocabulary splits) and train from scratch. We use the default set of hyper-parameters, and use 3 redundancy classifiers.

\section{\hlps: Instance Segmentation}

\subsection{Segmentation Tree Generation}\label{sec:sup:tree_gen}
Given an input point cloud, we first make a network pass and classify points into \thing, \stuff, and \other classes. Then, we construct a hierarchical tree segmentation tree $T$ by applying HDBSCAN~\cite{Mcinnes2017OSS} on points. 
Concretely, at each level of the hierarchy, we reduce the distance threshold $\epsilon$ (HDBSCAN connectivity hyperparameter) to obtain finer point segments. Therefore, the nodes in $T$ contain strictly smaller and finer-grained instances in child nodes. We follow~\cite{hu2020learning} and use distance thresholds $\epsilon \in [1.2488, 0.8136, 0.6952, 0.594, 0.4353, 0.3221]$.

\subsection{Learning Objectness-Scoring Function}

Given a hierarchical tree $T$ over the input point cloud, we need to find a node partitioning such that each point is assigned to a unique instance. Naturally, some nodes in the tree would contain high-quality segments, while others would consist of a soup of segments or overly segmented instances.
To associate a metric quality for each node, we follow \cite{hu2020learning} and learn a function to score each segment in the node. 

\PAR{Network Architecture.} Each node in the tree consists of a segment (\ie, a group of points), where the number of points may vary. We concatenate all the points in a segment to get a $N \times 3$ dimensional tensor, where $N$ is the number of points in the segment. There are several ways how to learn such a function $f(p) \to [0,1]$ that estimates how likely a subset of points represents an object. 
One approach is to estimate a per-point objectness score. Following~\cite{aygun21cvpr}, this can be learned by regressing a truncated distance $O \in \mathbb{R}^{N \times 1}$ to the nearest point center of a labeled instance~\cite{aygun21cvpr} on-top of decoder features $F \in \mathbb{R}^{N \times D}$. The objectness value can then be averaged over the segment $p \subset P$. 

Alternatively, we can train a holistic classifier as a second-stage network.
 The network comprises three major components: a) input projection layer, b) segment embedding layer, and c) objectness head. In the input projection layer, we project the input point cloud to a higher dimension of $N \times 256$ by passing through two fully-connected layers. 
Then, we compute a per-segment embedding of dimension $512$ using the embedding layer. This consists of set abstraction layers inspired by PointNet++~\cite{Qi17NIPS}, followed by a reduction over the points.

\PAR{Network Training.} The objectness head predicts per-segment objectness, using three fully-connected layers with a hidden layer of size 256. To obtain training supervision, we pre-built hierarchical segmentation trees $T_i$ for each point cloud $i$ in the training set and minimize the training loss based on the signal we obtain from \textit{matched} segments between the segmentation trees and set of labeled instances, $GT_i$. As described in Sec 4 of the main paper, for each node in the segmentation tree, the regressor predicts the objectness value  which is supervised by the intersection-over-union of the segment with the maximal matching ground-truth instance. 

%
Alternatively, we can also formulate the network as a classifier, where the post-softmax outputs from the network can be viewed as the quality of each segment (\ie, how good or bad a segment is), and this is trained using a binary cross-entropy loss.
We observe that the regression formulation empirically results in a better tree cut compared to the classifier formulation. 
We attribute this to the over-confident and peaky distributions resulting from the classifier.
The regressor formulation benefits from a smoother distribution over objectness scores, resulting in a better tree cut. We evaluate the aforementioned variants in Tab. 1 in the main paper.

\subsection{Inference}
With a score assigned to each segment, we now need to find a global segmentation, \ie, the optimal instance segmentation from an exponentially large space of possible segments. The global segmentation score is defined as the worst objectness among the individual segments in a tree cut. The optimal partition is the one that maximizes the global segmentation score. 
We outline this inference algorithm in Alg.~\ref{alg:huseg}. Each node in the tree $T$ constitutes a segment proposal for an object instance. For each proposal $S$, we score its objectness using the learned objectness function $f$. The segment $S$ is deemed as an optimal node to perform a tree-cut if its objectness is greater than any of its child nodes. By design, this tree-cut algorithm ensures that each segment is assigned to a unique instance.
For details about the algorithm and optimality guarantees, we refer the reader to~\cite{hu2020learning}.

\begin{algorithm}
\caption{Node Partitioning given a Hierarchical Segmentation Tree}\label{alg:huseg}
\begin{algorithmic}[1]
    \Function{Tree-Cut}{S: Point Segment, T: Tree, f($\cdot$): Scoring Function} 
    \State $F_S \gets f(S)$ \Comment{Predicted objectness for segment S}
    \State $C_S \gets$ Set of children nodes for $S$ in $T$
    \For{$S_i$ in $C_S$}
        \State $T_i \gets$ subtree of $T$ with $S_i$ as root
        \State $S_i, F_i \gets$ \Call{Tree-Cut}{$S_i, T_i, f$} \Comment{Score each segment in the child node}
        \If{$F_i \leq F_S$}
            \Return $S, F_S$
        \EndIf
    \EndFor
    \If{$min_i F_i > F_S$}
        \State $S \gets \bigcup_i S_i$
        \State $F_S \gets \min_i F_i$
    \EndIf        
    \Return $S, F_S$
    \EndFunction
\end{algorithmic}

\end{algorithm}

\section{Implementation Details}

\PAR{Training encoder-decoder network.} To train the point-classification network on each vocabulary, we follow the training procedure from~\cite{aygun21cvpr}. We train the network for 1000 epochs with a batch size of 8. We use the SGD optimizer with a learning rate of $1e-3$ and a linear decay schedule. 

\PAR{Second-stage training.} In contrast to the encoder-decoder network, which takes an entire point cloud as input, the second-stage network requires positive and negative training instances (\ie, examples of objects and non-objects) from the segmentation tree to evaluate the loss functions (regression or cross-entropy losses). To generate these instances, we first generate semantic predictions. Next, we use the points classified as \thing or \other to generate the segmentation tree using thresholds as described in Sec~\ref{sec:sup:tree_gen}. Each node in the segmentation tree is a training sample for the second-stage network. We use predictions from the encoder-decoder instead of ground-truth semantic labels, since during inference the second-stage network must be robust to misclassification errors within each node in the segmentation tree. 

\PAR{Training objectness regression function.} To train the regressor, we need to generate the corresponding ground-truth for each segment in the generated training set.
For a given segment, the target score is computed as the maximum intersection-over-union of the segment with all the ground-truth instances in the dataset. Finally, we train the network using a mean-squared error loss function with a learning rate of $2e-3$ and batch size of 512 for 200 epochs. 

\PAR{Training objectness classification function.} Alternatively, learning an objectness function can be posed as a classification problem, rather than a regression problem. In this case, we supervise the network via cross-entropy loss. 
The target labels for training this classifier are obtained by binarizing the regression targets using pre-defined intersection-over-union (IoU) thresholds. 
A regression target with an IoU greater than $0.7$ is defined as a positive segment, and a regression target with an IoU less than $0.3$ is treated as a negative segment. 
Since the ground-truth classification targets are generated from the hierarchical tree, which consists of predicted {\tt known-things} or \other, the generated training data is strongly biased towards positive samples. To elaborate, since the point classification network performs well, the segments in the tree most likely consist of  {\tt known-things} or \other, barring misclassification error. Therefore, while training this network, we observe a disproportionate imbalance towards the positive class. To mitigate this, perform a weighted resampling; We resample instances (segments) of either positive or negative classes with a probability proportional to the inverse frequency of that class.

\section{Complexity Analysis}
\label{sec:sup:complexity}

\revision{
\hlps requires a two-stage training process. The first stage method, 4DPLS~\cite{aygun21cvpr} is not a real-time. In addition, the second stage requires the construction of a segmentation tree. Given $N$ points, the algorithm's~\cite{hu2020learning} time complexity and space complexity are linear in $N$. In practice, we observe that $N$ is quite large, often of the order of 100,000+ points per scan. Therefore, a limitation of \hlps is that it cannot be run in real-time.
}






\end{appendices}
\clearpage

\bibliographystyle{sn-mathphys}      
\bibliography{main}


\end{document}